\documentclass[12pt]{l4dc2023}
\usepackage[mathscr]{euscript}
\usepackage{times}

\newcommand{\diag}{\ensuremath{\operatorname{diag}}}
\newcommand{\bs}{\ensuremath{\mathbf{s}}}
\newcommand{\bu}{\ensuremath{\mathbf{u}}}
\newcommand{\bj}{\ensuremath{\mathbf{j}}}
\newcommand{\bk}{\ensuremath{\mathbf{k}}}
\newcommand{\bell}{\ensuremath{\boldsymbol{\ell}}}
\newcommand{\bbf}{\ensuremath{\mathbf{f}}}
\newcommand{\bg}{\ensuremath{\mathbf{g}}}
\newcommand{\bX}{\ensuremath{\mathbf{X}}}
\newcommand{\bL}{\ensuremath{\mathbf{L}}}

\newcommand{\bx}{\ensuremath{\mathbf{x}}}
\newcommand{\by}{\ensuremath{\mathbf{y}}}

\newcommand{\bpsi}{\ensuremath{\boldsymbol{\psi}}}
\newcommand{\blambda}{\ensuremath{\boldsymbol{\lambda}}}

\newcommand{\btheta}{\ensuremath{\boldsymbol{\theta}}}

\newcommand{\bbR}{\ensuremath{\mathbb{R}}}

\title[L\'{e}vy $\alpha$-Stable Drift Identification]{Drift Identification for L\'{e}vy alpha-Stable Stochastic Systems}

\author{%
\Name{Harish~S. Bhat} \Email{hbhat@ucmerced.edu}\\
\addr Department of Applied Mathematics, University of California, Merced, 5200 N. Lake Rd, Merced, CA 95343
}

\begin{document}
\maketitle

%%%%%%%%%%%%%%%%%%%%%%%%%%%%%%%%%%%%%%%%%%%%%%%%%%%%%%%%%%%%%%%%%%%%%%%%%%%%%%%%
\begin{abstract}
This paper focuses on a stochastic system identification problem: given time series observations of a stochastic differential equation (SDE) driven by L\'{e}vy $\alpha$-stable noise, estimate the SDE's drift field.  For $\alpha$ in the interval $[1,2)$, the noise is heavy-tailed, leading to computational difficulties for methods that compute transition densities and/or likelihoods in physical space.  We propose a Fourier space approach that centers on computing time-dependent characteristic functions, i.e., Fourier transforms of time-dependent densities.  Parameterizing the unknown drift field using Fourier series, we formulate a loss consisting of the squared error between predicted and empirical characteristic functions.  We minimize this loss with gradients computed via the adjoint method.  For a variety of one- and two-dimensional problems, we demonstrate that this method is capable of learning drift fields in qualitative and/or quantitative agreement with ground truth fields.
\end{abstract}

%%%%%%%%%%%%%%%%%%%%%%%%%%%%%%%%%%%%%%%%%%%%%%%%%%%%%%%%%%%%%%%%%%%%%%%%%%%%%%%%
\section{Introduction}
\label{sect:intro}
Consider a stochastic differential equation driven by L\'{e}vy $\alpha$-stable noise:
\begin{equation}
\label{eqn:sde}
\mathrm{d} \bX_t = \bbf(\bX_t) \mathrm{d}t + G(\bX_t) \mathrm{d} \bL_t.
\end{equation}
Here $\bX_t$ is an $\bbR^d$-valued stochastic process, $\bbf : \bbR^d \to \bbR^d$ is the drift function (a vector field), $G : \bbR^d \to \bbR^{d \times d}$ is the diffusion function (a matrix field), and $\bL_t$ is a L\'{e}vy symmetric $\alpha$-stable process in $\bbR^d$.  We define this process in detail below, but for the sake of intuition, we mention two special cases.  When $\alpha=1$, the $\alpha$-stable density is the Cauchy density, $p(x) = (\pi(1+x^2))^{-1}$.  When $\alpha=2$, the $\alpha$-stable density is Gaussian.  We focus entirely on $\alpha \in [1, 2)$, in which case increments of $\bL_t$ are heavy-tailed and have infinite variance.

Assume we have access to $n_T$ trajectories, each of which comprises observations of the state of (\ref{eqn:sde}) at discrete times $\{j \Delta t\}_{j=0}^N$.  Denote the $k$-th such trajectory by
$\bX^k = \{ {\bX}_0^k, {\bX}_1^k, \ldots, {\bX}_N^k \}$; then our data is $\mathcal{D} = \{ \bX^k\}_{k=1}^{n_T}$.    Assume that the diffusion function $G$ is given.
\emph{The system identification (or learning/inverse) problem we address is: Given trajectory data $\mathcal{D}$, estimate the drift $\bbf$ in (\ref{eqn:sde}).}

We consider (\ref{eqn:sde}) because of its capacity to model systems in biology \citep{ReynoldsFrye2007,PhysRevE.82.011121,ohta2022intrinsic}, engineering \citep{pmlr-v97-simsekli19a,tajmirriahi2021modeling,gan2021levy}, and finance \citep{HURST1999105,CarrWu2003,cartea2009option} that all feature heavy tails.  Despite the increasing importance of L\'{e}vy processes in modeling complex phenomena, \emph{system identification} for L\'{e}vy-driven SDE such as (\ref{eqn:sde}) has received far less attention than the corresponding problem for SDE driven by standard Brownian motion.

We parameterize $\bbf$ using parameters $\btheta$.  Suppose we try to estimate $\btheta$ by maximizing the likelihood $p( \mathcal{D} \, | \, \btheta)$.  To do this directly, we would need the SDE's transition density $p(\bX^k_{n+1} \, | \, \bX^k_n, \btheta)$, on a large enough spatial domain to capture our observations.  While it is possible to numerically solve a Fokker-Planck-Kolmogorov-type equation to obtain transition densities for particular initial conditions \citep{GAO20161}, it is not clear how to scale such methods to solve estimation/inference problems.  Additionally, the numerical experiments of \cite{fang2022end} indicate that even when the transition density is known, maximizing the likelihood directly does not yield accurate estimates of the drift and/or diffusion functions.

\emph{In this paper, we develop a simulation and estimation framework for (\ref{eqn:sde}) that resides entirely in Fourier space.}  When $\alpha=1$, the $\alpha$-stable characteristic function (i.e., Fourier transform of the density) is $\psi(s) = \exp(-|s|)$.  A small grid in $s$ space is sufficient to represent $\psi(s)$ accurately.  In contrast, to represent the Cauchy density $p(x)$ accurately, we need a large grid in $x$ space, especially if we wish to preserve the normalization $\int_{\bbR} p(x) \, \mathrm{d}x = 1$.  More generally, for $\alpha < 2$, because increments of the process $\bL_t$ have heavy tails, we expect densities of the solution $\bX_t$ of (\ref{eqn:sde}) to decay to zero \emph{slowly} as $\|\bx\| \to \infty$.  By Fourier duality, if the density is delocalized in space, then its Fourier transform must be localized in Fourier space \cite[Chap. 2]{mallat2008wavelet}.  \emph{By staying in Fourier space, we aim for both computational tractability and accurate simulation/estimation.}

Let us review prior studies that address system identification problems for (\ref{eqn:sde}).  In the \emph{parametric} version of the problem, the drift $\bbf$ and diffusion $G$ are specified up to a finite-dimensional set of parameters that one seeks to learn.  \cite{hongwei2010parameter} and \cite{LONG20171475} address the parametric problem using least squares estimators, while \cite{Jasra2019} pursues a Bayesian approach.

% For instance, one might have $f(x; \theta) = \theta x$ and $G \equiv 1$; the task is then to estimate $\theta$ from data.  

In the \emph{nonparametric} problem, the goal is to learn the functional form of $\bbf$, and possibly that of $G$ as well.  Early work on this problem focuses on the high-frequency setting in which the time between observations tends to zero \citep{schmisser2014non}. For versions of the problem in which data is sampled regularly in time, a variety of modern methods have recently been applied, including Koopman operators \citep{Lu2020}, nonlocal Kramers-Moyal expansions \citep{Li2021,li2022extracting}, normalizing flows \citep{Lu2022,li2021extracting}, minimization of a Kullback-Leibler loss with Fokker-Planck constraints \citep{Duan2022Chaos}, and neural networks \citep{Chen2021,Wang2022,fang2022end}.  These approaches all work in physical space, necessitating approximations and/or ad hoc techniques to render the problem tractable.  Note that time-dependent characteristic functions have been used to understand problems \emph{other} than system identification for L\'{e}vy-driven SDE.  We review this literature in Section \ref{sect:litreview} of the Appendix.

To our knowledge, no prior work on system identification considers recasting in Fourier space the equations of motion, the loss function, and/or the representation of unknown fields.  In the present work, we show that a relatively unsophisticated Fourier space method can still successfully identify drifts in SDE driven by Cauchy noise, e.g., (\ref{eqn:sde}) with $\alpha=1$.  For periodic vector fields, we obtain quantitative agreement between learned and true drifts. For non-periodic polynomial vector fields, we obtain quantitative agreement in $d=1$ and qualitative agreement in $d=2$. Taken together, these results approach the limits of prior methods.

%%%%%%%%%%%%%%%%%%%%%%%%%%%%%%%%%%%%%%%%%%%%%%%%%%%%%%%%%%%%%%%%%%%%%%%%%%%%%%%%
\section{Forward Problem}
\label{sect:forwardprob}
  Throughout this work, we use boldface lowercase letters for vectors/vector fields; boldface uppercase letters for vector-valued random variables and stochastic processes; and standard uppercase letters for matrices.  We think of vectors $\mathbf{a}, \mathbf{b} \in \mathbb{R}^d$ as columns, so that $\mathbf{a}^T \mathbf{b}$ is a scalar; here $^T$ denotes transpose.
  
Let $p(\bx,t)$ denote the probability density function (PDF) of the exact solution $\bX_t$ of (\ref{eqn:sde}).  Then the forward problem is: \emph{Given $p(\bx,0)$, compute $p(\bx,t)$ for $t > 0$.}  To solve the forward problem, we must introduce a few mathematical notions.  Given any $\bbR^d$-valued random variable $\bX$ with density $p(\bx)$, we can define the characteristic function as the Fourier transform of the density:
\begin{equation}
\label{eqn:forft}
\psi(\bs) = \widehat{p}(\bs) = E[e^{i \bs^T \bX}] = \int_{\bx \in  \bbR^d} e^{i \bs^T \bx} p(\bx) \, \mathrm{d}\bx.
\end{equation}
A random variable is determined by its characteristic function, and characteristic functions (unlike, e.g., moment-generating functions) can be used to distinguish numerically between random variables with different distributions \citep{mccullagh1994does, waller1995does, luceno1997further}.
Note that
\begin{equation}
\label{eqn:normalization}
\psi(\mathbf{0}) = \int_{\bx \in \bbR^d} p(\bx) \, \mathrm{d}\bx = 1.
\end{equation}
Returning to definitions: let $\mathrm{L}_t^j$ denote the $j$-th component of $\bL_t \in \bbR^d$.  Here we assume that, for $j \neq k$, $\mathrm{L}_t^j$ and $\mathrm{L}_t^k$ are independent scalar $\alpha$-stable L\'{e}vy processes defined by the following criteria: (i) $\mathrm{L}_0^j = 0$ almost surely,
(ii) $\mathrm{L}_t^j$ has independent increments, and (iii) For $t_2 > t_1 \geq 0$, $\mathrm{L}_{t_2}^j - \mathrm{L}_{t_1}^j$, the increment over a time interval of length $t_2 - t_1$,  has an $\alpha$-stable distribution with scale parameter $\sigma = (t_2 - t_1)^{1/\alpha}$, skewness parameter $\beta=0$, and location parameter $\mu=0$.

The term $d\bL_t$ in SDE (\ref{eqn:sde}) is shorthand for an increment of the $\bL_t$ process over a time interval $\mathrm{d}t$; for further details, consult  \cite{applebaum2009levy}.  For our purposes, it is sufficient to work with the Euler-Maruyama temporal discretization of (\ref{eqn:sde}) \citep{janicki2021simulation}:
\begin{equation}
\label{eqn:em}
\bx_{n+1} = \bx_n + \bbf(\bx_{n}) h + G(\bx_{n}) \Delta \bL_{n+1}.
\end{equation}
The random vector $\Delta \bL_{n+1}$, which is independent of $\bx_n$, consists of independent increments of $\alpha$-stable processes all over a time interval $h$.  Therefore, its characteristic function is
\begin{equation}
\label{eqn:emchar}
\psi_{\Delta \bL_{n+1}}(\bs) = \prod_{j=1}^d \exp(- h |s_j|^\alpha).
\end{equation}
Let $\widetilde{p}(\bx,t_{n+1})$ denote the PDF of $\bx_{n+1}$. Let us denote the conditional density of $\bx_{n+1}$ given $\bx_n = \by$ by $p_{n+1, n}(\bx | \by)$.  Marginalization yields an evolution equation for the marginal density of $\bx_n$:
\begin{equation}
\label{dtq}
\widetilde{p}(\bx, t_{n+1})=\int_{\by \in \bbR^d} p_{n+1, n}(\bx | \by ) \widetilde{p}(\by, t_n) \, \, \mathrm{d} \by.
\end{equation}
Assume that $G(\by) = \diag \bg(y)$. Computing the characteristic function of both sides, we obtain
\begin{equation}
\label{eqn:CFupdate}
\psi_{n+1}(\bs) = \int_{\by \in \bbR^d }e^{i \bs^T \left(\by + \bbf(\by)h\right) -h |\bs^T \bg(\by)|^\alpha }\widetilde{p}(\by, t_n) \, \mathrm{d}\by.
\end{equation}
To derive this, we start from (\ref{eqn:emchar}) and build up the right-hand side of (\ref{eqn:em}), conditional on $\bx_n = \by$. Scaling $\Delta \bL_{n+1}$ by $G(\by)= \diag \bg(y)$ yields the characteristic function $\exp(-h |\bs^T \bg(\by)|^\alpha)$. Translation by $\by + \bbf(\by) h$ yields the prefactor $\exp(i \bs^T (\by + \bbf(\by) h))$.  Putting these pieces together gives (\ref{eqn:CFupdate}).  Applying the inverse transform $\widetilde{p}(\by,t_n) = (2 \pi)^{-d} \int_{\bu \in \bbR^d} e^{-i \bu^T \by} \psi_n(\bu) \, \mathrm{d}\bs$, we obtain
\begin{subequations}
\begin{align}
\label{eqn:ctq}
\psi_{n+1}(\bs) &=  \int_{\bu \in \bbR^d} \widetilde{K}(\bs,\bu)\psi_{n}(\bu) \, \mathrm{d}\bu \\
\label{eqn:kdef}
\widetilde{K}(\bs,\bu) &= \frac{1}{(2\pi)^d}\int_{\by \in \bbR^d }e^{i \bs^T \left(\by+\bbf(\by)h\right) -h |\bs^T \bg(\by)|^\alpha-i \bu^T \by}\, \mathrm{d} \by.
\end{align}
\end{subequations}
It may seem as though we can directly apply quadrature to evaluate (\ref{eqn:kdef}) and thereby obtain numerical methods to track the time-evolution of the characteristic function.   However, note that in the $h \to 0$ limit, the kernel $\widetilde{K}(\bs,\bu)$ becomes $\delta(\bs-\bu)$.  For $h > 0$ sufficiently small, typical quadrature methods cannot handle the nearly singular nature of (\ref{eqn:kdef}).  To cope, we Taylor expand $\widetilde{K}$ in the time step $h$. We present the derivation for the special case when $\bg(\by) = \bg$, a constant vector:
\begin{align}
\widetilde{K}(\bs,\bu) &= \frac{1}{(2\pi)^d} \! \! \int\displaylimits_{\by \in \bbR^d} \! \! e^{i (\bs - \bu)^T \by} e^{h (i \bs^T \bbf(\by) -|\bs^T \bg(\by)|^\alpha) } \, \mathrm{d}\by = \frac{e^{ -h|\bs^T \bg|^\alpha }}{(2\pi)^d}  \int_{\by \in \bbR^d} e^{i (\bs - \bu)^T y} e^{h i \bs^T \bbf(\by) } \, \mathrm{d}\by \nonumber \\
 &= \frac{1}{(2\pi)^d} e^{ -h|\bs^T \bg|^\alpha } \int_{\by \in \bbR^d} e^{i (\bs - \bu)^T \by} \biggl[ 1 + h i \bs^T \bbf(\by) - \frac{h^2}{2} (\bs^T \bbf(\by))^2 + O(h^3) \biggr] \, \mathrm{d}\by \nonumber \\
 \label{eqn:ourke}
 &\approx  e^{ -h|\bs^T \bg|^\alpha } \biggl[ \delta(\bs-\bu) + \frac{ h i }{(2 \pi)^d} \bs^T \widehat{\bbf}(\bs-\bu)  - \frac{h^2}{2 (2 \pi)^d} \bs^T \widehat{\bbf \bbf^T}(\bs-\bu) \bs \biggr] 
\end{align}
To justify truncation of the above expansion at second-order in $h$, we note that the Euler-Maruyama discretization (\ref{eqn:em}) itself has $O(h^2)$ local truncation error.  For examples of characteristic function evolution (\ref{eqn:ctq}) and kernel expansion (\ref{eqn:ourke}), see Section \ref{sect:properties} of the Appendix.

\emph{The expansion (\ref{eqn:ourke}) extracts and exposes the $\delta$ singularity.}  As we will only use (\ref{eqn:ourke}) under the integral in (\ref{eqn:ctq}), the integral against $\delta(\bs-\bu)$ yields a contribution of $\psi_n(\bs)$.  With the $\delta$ singularity out of the way, we can treat the rest of the integral in a number of ways.

While the derivation can easily be extended to the case where $\bg(\by)$ is not constant in $\by$, we leave a complete exploration of that version of the method (including numerical results) for future work.  For the remainder of the present work, we assume that $\bg$ is a known, constant vector.  We focus on applying (\ref{eqn:ourke}) and (\ref{eqn:ctq}) to the problem of identifying the drift $\bbf$.  Note that combining (\ref{eqn:ourke}) and (\ref{eqn:ctq}) yields the following discrete-time, continuous-space \emph{characteristic function evolution} equation: 
\begin{equation}
\label{eqn:charfunevol}
\psi_{n+1}(\bs) = e^{-h | \bs^T \bg |^\alpha} \biggl[ \psi_n(\bs) 
+ \! \!  \! \int\displaylimits_{\bu \in \mathbb{R}^d} \! \! \biggl( \frac{ h i }{(2 \pi)^d} \bs^T \widehat{\bbf}(\bs-\bu) \, \psi_n(\bu) - \frac{h^2}{2 (2 \pi)^d} \bs^T \widehat{\bbf \bbf^T}(\bs-\bu) \bs \, \psi_n(\bu)  \biggr) \mathrm{d} \bu \biggr].
\end{equation}

\section{System Identification}
\label{sect:inverse}
Here we address the system identification problem described in Section \ref{sect:intro}.
To begin, assume that $\bbf$ is parameterized by $\btheta$---we detail this below.  We will estimate $\btheta$ by minimizing a loss function that consists  of the squared difference between predicted and empirical characteristic functions.  To unpack this, first note that that each trajectory gives rise to a time-dependent empirical density consisting of a sequence of point masses: for $j = 0, 1, \ldots, N$, $\widetilde{p}^k(\mathbf{x}, t_j) =  \delta(\mathbf{x} - \mathbf{X}^k_j)$.    Averaging over all $n_T$ trajectories and applying the Fourier transform (\ref{eqn:forft}), we obtain the \emph{empirical characteristic function} of our collection of trajectories: for $j = 0, 1, \ldots, N$,
\begin{equation}
\label{eqn:empiricalcharfun}
\widetilde{\psi}(\mathbf{s}, t_j) = \frac{1}{n_T} \sum_{k=1}^{n_T} \exp(i \mathbf{s}^T \mathbf{X}^k_j).
\end{equation}
Now fix $j \in [0, \ldots, N-1]$.  Let $h = \Delta t / \nu$ for some integer $\nu \geq 1$ sufficiently large so that (\ref{eqn:charfunevol}) is stable and accurate. Take (\ref{eqn:empiricalcharfun}) as the \emph{initial condition} $\psi_0(\bs)$ for the evolution equation (\ref{eqn:charfunevol}).  Given parameters $\btheta$ that determine $\bbf$, we iterate (\ref{eqn:charfunevol}) for $\nu$ steps.  We obtain a \emph{predicted} characteristic function corresponding to time $t_{j+1}$ in our data set---let us denote it as $\psi(\bs, t_{j+1}; \btheta)$.  We then form 
\begin{equation}
\label{eqn:mmdloss}
\Lambda_{\mathscr{C}}(\btheta) = \frac{1}{2} \sum_{j=0}^{N-1} \int\limits_{\bs \in \mathbb{R}^d} \Bigl\| \psi(\bs, t_{j+1}; \btheta) - \widetilde{\psi}(\bs, t_{j+1}) \Bigr\|^2 \, \mathrm{d} \bs.
\end{equation}
A key feature of the loss $\Lambda_{\mathscr{C}}(\btheta)$ is that it does not require the evaluation of any probability density functions in real space.  To put it another way, we can compute (\ref{eqn:mmdloss}) without computing the inverse Fourier transforms of our predicted characteristic functions.  This is contrast to, for instance, loss functions based on the negative log likelihood. Especially for $\alpha$ near $1$, we expect that predicted densities (equivalently, inverse Fourier transforms of predicted characteristic functions) will require massive spatial domains to capture heavy-tailed behavior.  We prefer to stay in Fourier space.

Note that the loss (\ref{eqn:mmdloss}) is a special case of the \emph{maximum mean discrepancy} (MMD) loss, which has been used in other areas of machine learning \citep{NIPS2015_b571ecea, MAL-060}.   We have not seen (\ref{eqn:mmdloss}) used before in system identification problems.

\paragraph{Representation/Parameterization of the Drift Field.}
Let $\phi_m$ denote the $m$-th component of a vector field $\phi$.  Assume there exists an integer $L > 0$ such that $\phi_m \in \mathscr{L}^2([-L \pi, L \pi]^d, \bbR)$ for all $m$.  Then the Fourier series expansion of $\phi$ converges in $\mathscr{L}^2$ to $\phi$.  By choosing $L$ sufficiently large, we can use Fourier series to represent many well-behaved $\mathscr{L}^2$ vector fields.

Let $J$ be a positive integer indicating how many Fourier modes we wish to use.  Let $\bj = (j_1, \ldots, j_d)$ be a multi-index, a vector of integers.  Let $\mathcal{J} = \{ \bj \in \mathbb{Z}^d \, | \, |j_{\ell}| \leq J \text{ for all $\ell$ }\}$.  Then, for the $m$-th component of $\bbf$, our Fourier series model and its Fourier transform are:
\begin{equation}
\label{eqn:L2drifthat}
f_m (\bx; \btheta) = \sum_{\bj \in \mathcal{J}} \theta^{\bj}_m e^{i \bj^T \bx / L} \ \Longrightarrow \
\widehat{f}_m (\bs) = (2 \pi)^d \sum_{\bj \in \mathcal{J}} \theta^{\bj}_m \delta(\bs + \bj/L).
\end{equation}
From (\ref{eqn:L2drifthat}), $f_{m} (\mathbf{x}) f_{m'} (\mathbf{x}) = \sum_{\bj, \bj'  \in \mathcal{J}} \theta^{\bj}_m \theta^{\bj'}_{m'} e^{i (\bj+ \bj')^T \mathbf{x}/L } = \sum_{\bk \in \mathcal{K}}  \sum_{\bj  \in \mathcal{J}}  \theta^{\bj}_m \theta^{\bk - \bj}_{m'}   e^{i \bk^T \mathbf{x} / L }$,
with $\bk = \bj + \bj'$ and $\mathcal{K} = \{ \bk \in \mathbb{Z}^d \, | \, |k_m| \leq 2 J \text{ for all } m \}$. Then the Fourier transform is
\begin{equation}
\label{eqn:L2drift2hat}
\widehat{ f_m f_{m'} }(\bs) = (2 \pi)^d \sum_{\bk \in \mathcal{K}} \left[ \sum_{\bj \in \mathcal{J}}  \theta^{\bj}_m \theta^{\bk - \bj}_{m'} \right] \delta(\bs + \bk / L)
\end{equation}
Using (\ref{eqn:L2drifthat}), (\ref{eqn:L2drift2hat}), and the definition $\bigl( \btheta \ast  \btheta^T \bigr)^{\bk} =  \sum_{\bj \in \mathcal{J}}  \btheta^{\bj} \bigl( \btheta^{\bk - \bj} \bigr)^T$, we obtain
\begin{equation}
\label{eqn:hatresults}
\widehat{\mathbf{f}} (\bs) = (2 \pi)^d \sum_{\bj \in \mathcal{J}} \btheta^{\bj} \delta(\bs + \bj/L), \quad \text{ and } \quad
\widehat{\mathbf{f} \mathbf{f}^T} (\bs) = (2 \pi)^d \sum_{\bk \in \mathcal{K}} \bigl( \btheta \ast  \btheta^T \bigr)^{\bk} \delta(\bs + \bk / L).
\end{equation}

\paragraph{Forward Propagation.}
Substituting (\ref{eqn:hatresults}) into (\ref{eqn:charfunevol}), we derive
\begin{equation}
\label{eqn:psiupdate}
\psi_{k+1}(\bs) = e^{ -h |\bs^T \bg|^\alpha } \biggl[ \psi_k(\bs) + i h \bs^T  \sum_{\bj \in \mathcal{J}} \btheta^{\bj} \psi_k(\bs + \bj/L) 
	- \frac{h^2}{2} \bs^T \Bigl(  \sum_{\bk \in \mathcal{K}} \bigl( \btheta \ast  \btheta^T \bigr)^{\bk}  \psi_k(\bs + \bk / L) \Bigr) \bs   \biggr].
\end{equation}
By choosing a Fourier representation of $\bbf$, we have ensured that the transforms $\widehat{\bbf}$ and $\widehat{\bbf \bbf^T}$ involve Dirac deltas, enabling the exact evaluation of the integrals in (\ref{eqn:charfunevol}).  Had we chosen other representations of $\bbf$---e.g., polynomials, splines, or neural networks---we would have had to evaluate the integrals in (\ref{eqn:charfunevol}) via numerical quadrature.  This and the $\mathscr{L}^2$ property mentioned above is why we employ a Fourier series representation of $\bbf$.

To finally bring (\ref{eqn:psiupdate}) into a form suitable for numerical implementation, we must discretize the spatial variable $\bs$.  Here we track the pointwise values of $\psi$ on an equispaced grid in $\bs$ space, as in a finite-difference method.  Based on the form of (\ref{eqn:psiupdate}), we choose a grid with spacing $\Delta s = 1/(n_L L)$ for some integer $n_L \geq 1$.  The grid itself can be described by the collection of points $\mathcal{M} = \{ \bj \Delta s \, | \, |j_{\ell}| \leq M \text{ for all } \ell \}$.  Then evaluating both sides of (\ref{eqn:psiupdate}) at an arbitrary grid point $\bj \Delta s \in \mathcal{M}$, we obtain the following fully discrete (in space and time) scheme:
\begin{multline}
\label{eqn:psiupdate2}
\psi_{k+1}(\bj \Delta s) = e^{ -h \Delta s |\bj^T \bg|^\alpha } 
\biggl[ \psi_k(\bj \Delta s) + i (h \Delta s) \bj^T \sum_{\bk \in \mathcal{J}} \btheta^{\bk} \psi_k( (\bj + \bk n_L) \Delta s)  \\
	- \frac{(h \Delta s)^2}{2} \bj^T \Bigl(  \sum_{\bk \in \mathcal{K}} \bigl( \btheta \ast  \btheta^T \bigr)^{\bk}  \psi_k(( \bj + \bk n_L) \Delta s) \Bigr) \bj   \biggr] 
\end{multline}
With the assumption that $\psi_k(\bell \Delta s) = 0$ for any $\bell \Delta s \notin \mathcal{M}$, this gives us a closed system of equations to evolve $\psi_k$ forward in time, on the spatial grid $\mathcal{M}$, starting from an initial condition $\psi_0$.

Note that (\ref{eqn:psiupdate2}) automatically preserves the normalization of the densities associated to each characteristic function.  To see this, evaluate both sides of (\ref{eqn:psiupdate2}) at $\bj = \mathbf{0}$ to obtain $\psi_{k+1}(\mathbf{0}) = \psi_k(\mathbf{0})$.  If the initial condition satisfies $\psi_0(\mathbf{0}) = 1$, then $\psi_k(\mathbf{0}) = 1$ for all $k \geq 0$. By (\ref{eqn:normalization}), the associated densities are all properly normalized.  For  additional remarks on the accuracy and stability of (\ref{eqn:psiupdate}), see Section \ref{sect:accstab} in the Appendix.

\paragraph{Adjoint Method.} Spatially discretizing  (\ref{eqn:mmdloss}) on the grid $\mathcal{M}$ described above, we derive
\begin{equation}
\label{eqn:discretemmdloss}
\Lambda(\btheta) = \frac{1}{2} \sum_{j=0}^{N-1} \sum_{\bj \in \mathcal{M}} \Bigl\| \psi(\bj \Delta s, t_{j+1}; \btheta) - \widetilde{\psi}(\bj \Delta s, t_{j+1}) \Bigr\|^2.
\end{equation}
We omit a factor of $(\Delta s)^d$ as it plays no role in what follows.  Our goal is to minimize the discrete-space, discrete-time loss  (\ref{eqn:discretemmdloss}) subject to the dynamics (\ref{eqn:psiupdate2}).  This is akin to an optimal control problem in which the drift vector field $\bbf$ (parameterized by $\btheta$) plays the role of the control.  In Section \ref{sect:adjdeets} of the Appendix, we detail an adjoint method to solve this dynamically constrained minimization problem.  The net result of this method is an efficient algorithm to compute $\nabla_{\btheta} \Lambda$.  

We have coded Python functions that implement both the MMD loss function (\ref{eqn:discretemmdloss}) and its gradient with respect to the parameters $\btheta$, computed via the adjoint method.  We pass these functions to SciPy's trust region optimizer, making use of the Symmetric Rank-1 (SR1) quasi-Newton Hessian approximation method \citep{Byrd1996}.  In all cases, we use an initial guess for $\btheta$ consisting of an array of zeros. 

%%%%%%%%%%%%%%%%%%%%%%%%%%%%%%%%%%%%%%%%%%%%%%%%%%%%%%%%%%%%%%%%%%%%%%%%%%%%%%%%

\section{Numerical Results}
To test the method's ability to identify systems driven by L\'{e}vy $\alpha$-stable noise, we conduct tests with synthetic data sets.  In each test, we use the Euler-Maruyama method to generate trajectories from systems with known drift and diffusion fields $\widetilde{\bbf}$ and $\bg$.  Using this data, we apply the characteristic function evolution and adjoint method described above to learn $\btheta$, Fourier coefficients of our estimated drift field $\bbf$.   We detail our error metrics below; in all cases, the idea is to compare the estimated $\bbf$ against the ground truth $\widetilde{\bbf}$.  We include particularly relevant implementation notes; for other implementation details, please consult Section \ref{sect:implementationdetails} in the Appendix.

\begin{figure}[t]
\begin{center}
\includegraphics[width=2in]{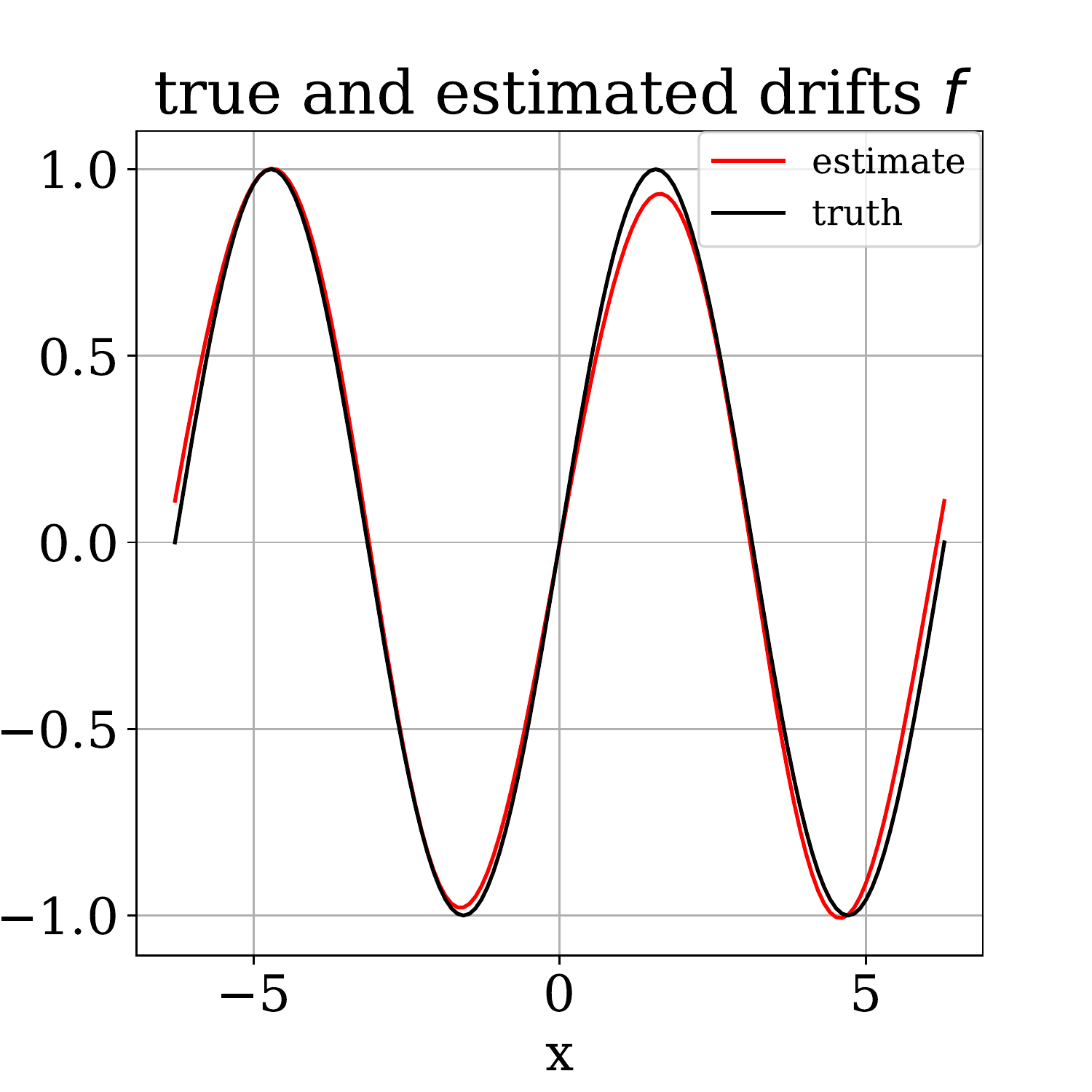} \includegraphics[width=1.875in]{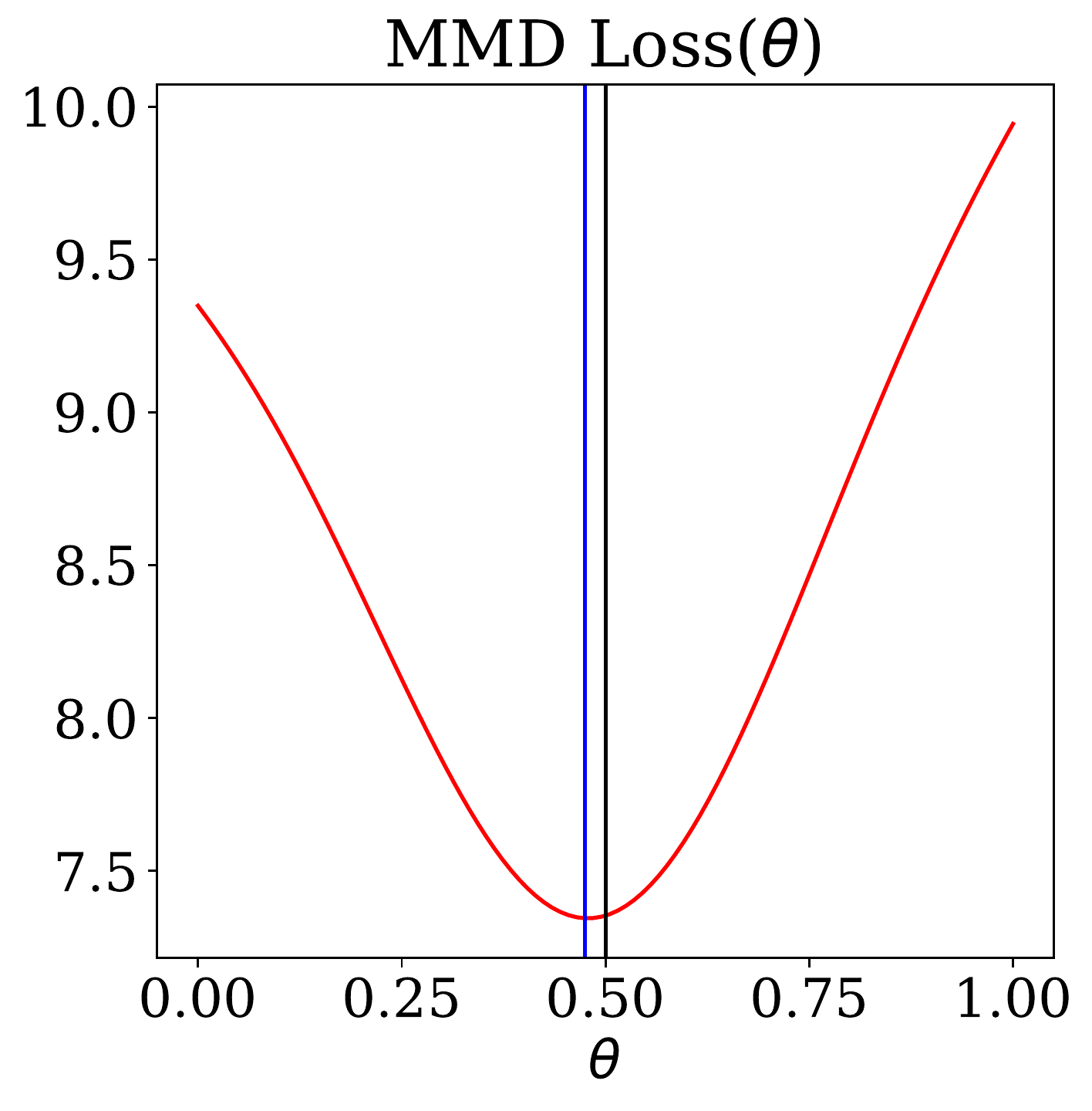}
\includegraphics[width=2in]{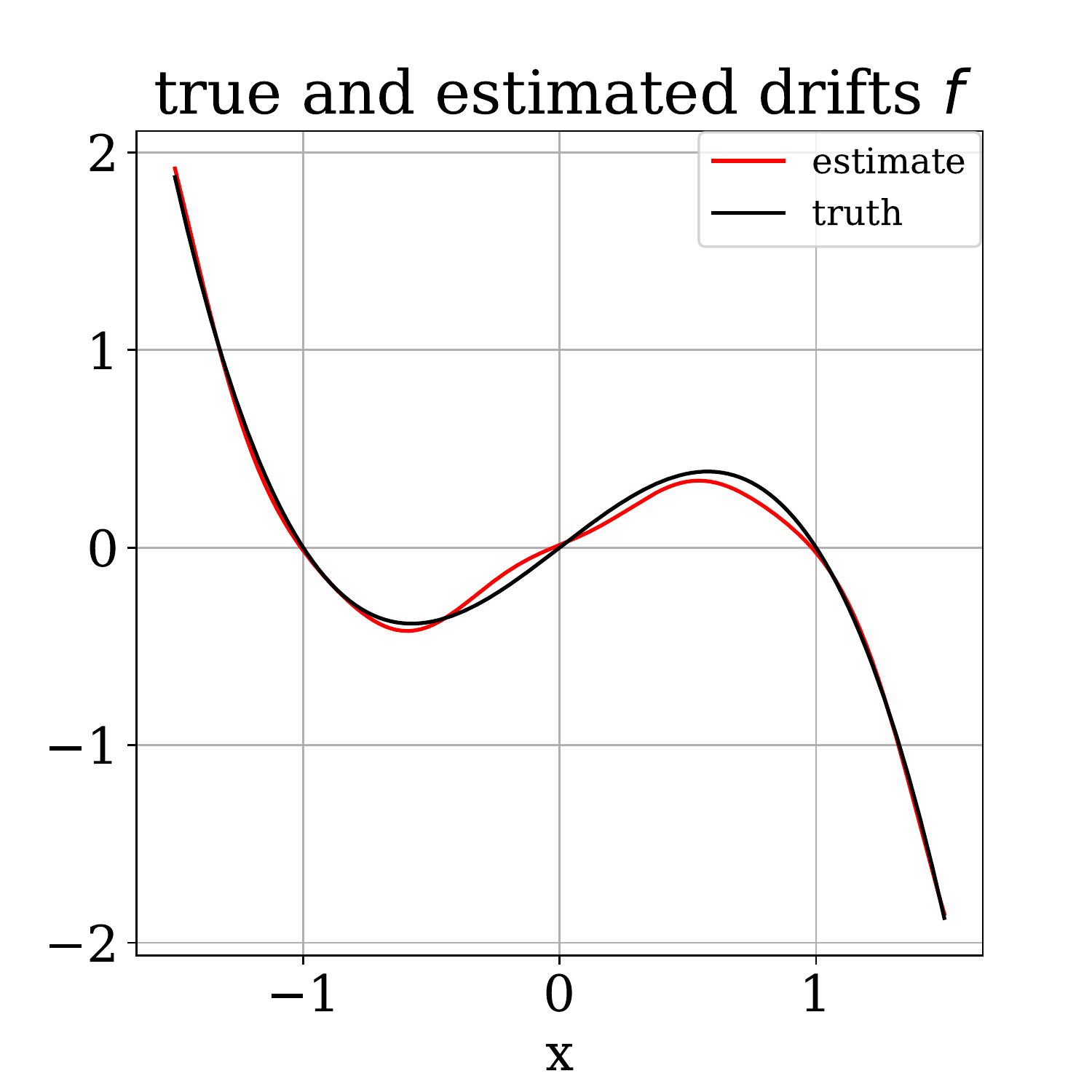}
\end{center}
\caption{We plot results for $\widetilde{f}(x) = \sin x$ (left, middle) and $\widetilde{f}(x) = x - x^3$ (right). In the left and right plots, we plot the learned $f$ (red) and true $\widetilde{f}$ (black).  Note the close agreement.}
\label{fig:onedimresults}
\end{figure}

\textbf{One-Dimensional Vector Fields ($d=1$).}
For all results in this section, we set $\alpha=1$, the most challenging case.  We set the diffusion constant to be $g = 0.25$.  All one-dimensional training data consists of $n_T = 100$ trajectories, each with initial condition $X_0 = 0$, $\Delta t = 10^{-1}$ and length $N = 41$.  We set $\nu = 100$ so that our method's internal time step is $h = \Delta t / \nu = 10^{-3}$.

For our first one-dimensional test, we generate data using the known vector field $\widetilde{f}(x) = \sin x$.   Using $J=4$ in our Fourier model, we apply the procedures from Section \ref{sect:inverse} to learn $\btheta \in \mathbb{C}^{2J+1}$.  In the left-most panel of Figure \ref{fig:onedimresults}, we plot $f$ and $\widetilde{f}$ in red and black, respectively, noting the close agreement between the two.  Let $\widetilde{\btheta}$ denote the Fourier coefficients for $\widetilde{f}$.   The mean absolute error (MAE) of $\btheta$ from $\widetilde{\btheta}$ is less than $3.2 \times 10^{-4}$.

To better understand the MMD loss, we set up an auxiliary model $\btheta = \mathbf{H}(\theta)$ for scalar $\theta$.  The purpose of $\mathbf{H}$ is to populate all entries of $\btheta$ with zeros except for those corresponding to the $j = \pm 2$ modes; we set the $j=2$ coefficient equal to $i \theta$ and the $j = -2$ coefficient equal to $-i \theta$.  Note that $\mathbf{H}(0.5) = \widetilde{\btheta}$, the ground truth Fourier coefficient vector.  In the middle panel of Figure \ref{fig:onedimresults}, we have plotted  $\Lambda(\mathbf{H}(\theta))$, the MMD loss (\ref{eqn:discretemmdloss}) evaluated on this auxiliary model that depends on only one real scalar.  \emph{Note that the MMD loss' global minimum (blue vertical bar) is quite close to the ground truth value (red vertical bar).}  This helps to justify our use of the MMD loss.

\begin{figure}[t]
\centering
\includegraphics[trim=65 0 65 0,clip,width=2.9in]{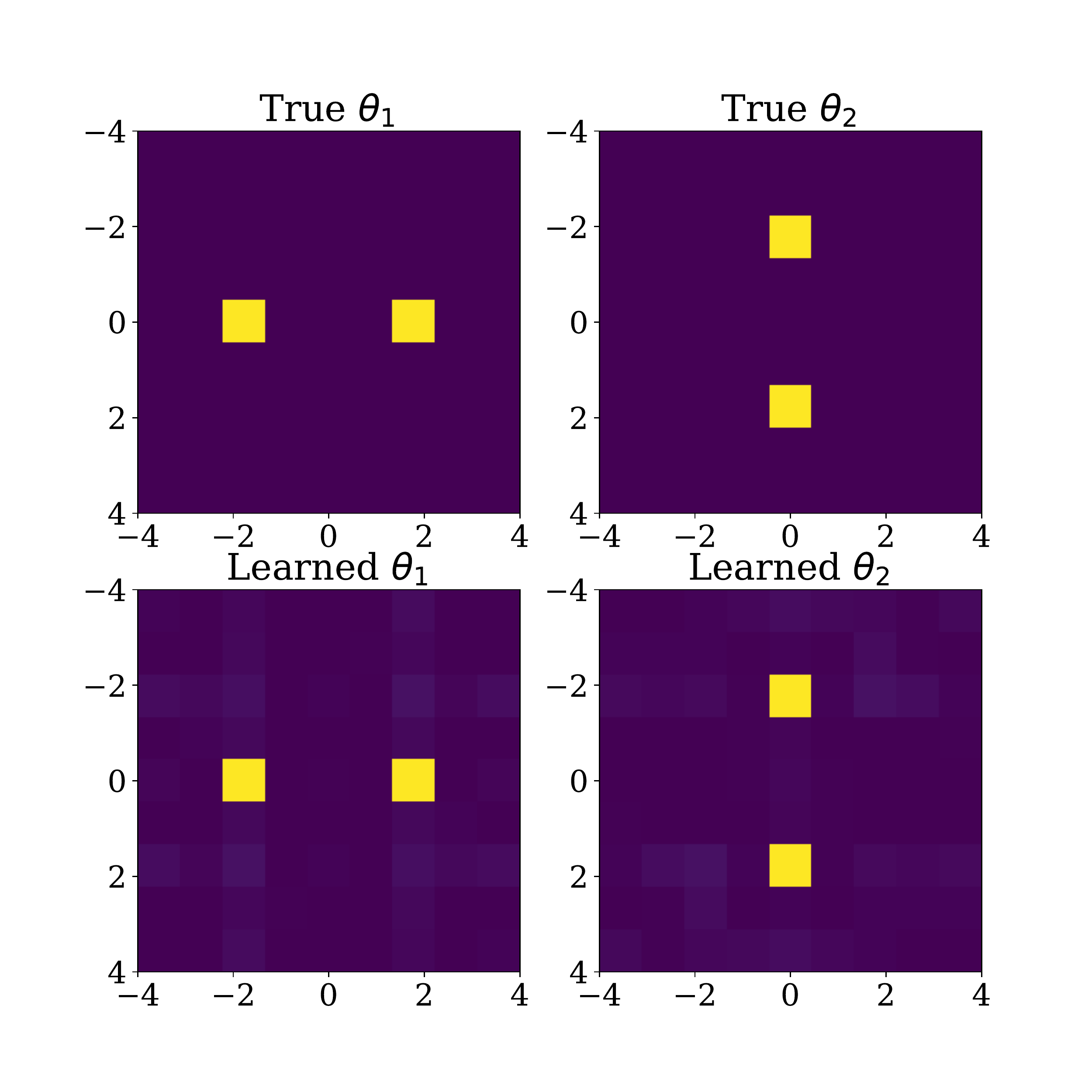} \ \vrule \ 
\includegraphics[trim=65 0 65 0,clip,width=2.9in]{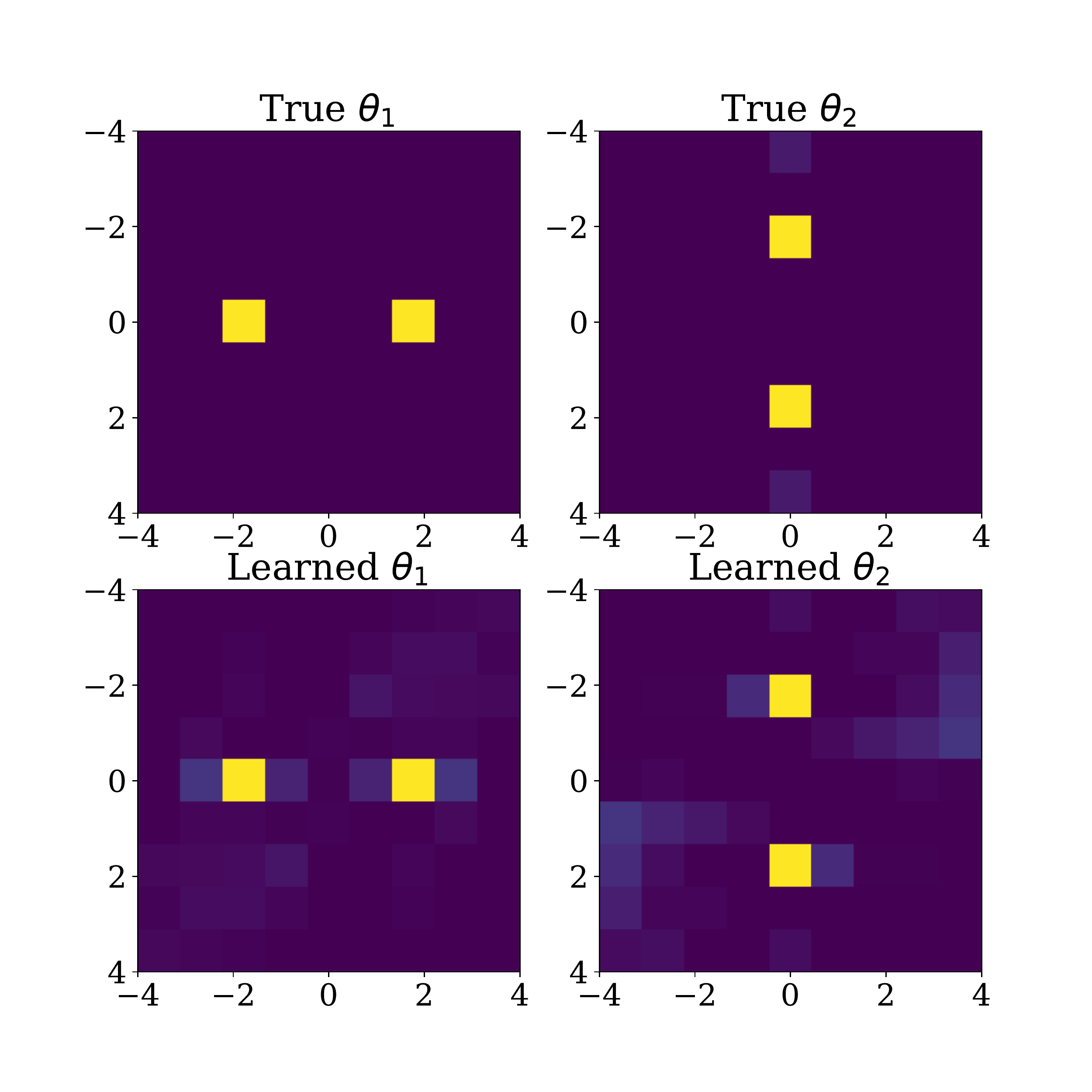} \caption{We plot true and learned results for  $\widetilde{\bbf}(x_1, x_2) = (\sin x_2, -\sin x_1)$ (single well potential, left of the vertical bar) and  $\widetilde{\bbf}(x_1, x_2) = (\sin x_2, -V'(x_1))$ with $V(x) = ((\sin x/2)^2 - 4)^2/10$ (double well potential, right of the vertical bar).  Each plot is a heatmap of $|\theta_q|$.}
\label{fig:twodimresults}
\end{figure}

Next, we consider  $\widetilde{f}(x) = x - x^3$, which is not periodic and does not have a sparse Fourier representation.  We choose this to simulate a real-world situation in which we do not have prior knowledge regarding the optimal basis to use in our model $f$.  With all hyperparameters as above, we learn  $\btheta \in \mathbb{C}^{2 J + 1}$ with $J = 16$.  In the right-most panel of Figure \ref{fig:onedimresults}, we plot $f$ and $\widetilde{f}$ in red and black, respectively.  We note that the curves agree closely on the domain $[-3/2, 3/2]$.  For further analysis of the errors in this case, please see Section \ref{sect:onedimdblwell} in the Appendix.

\begin{figure}[t]
\centering
\includegraphics[trim=10 0 50 0,clip,height=2.55in]{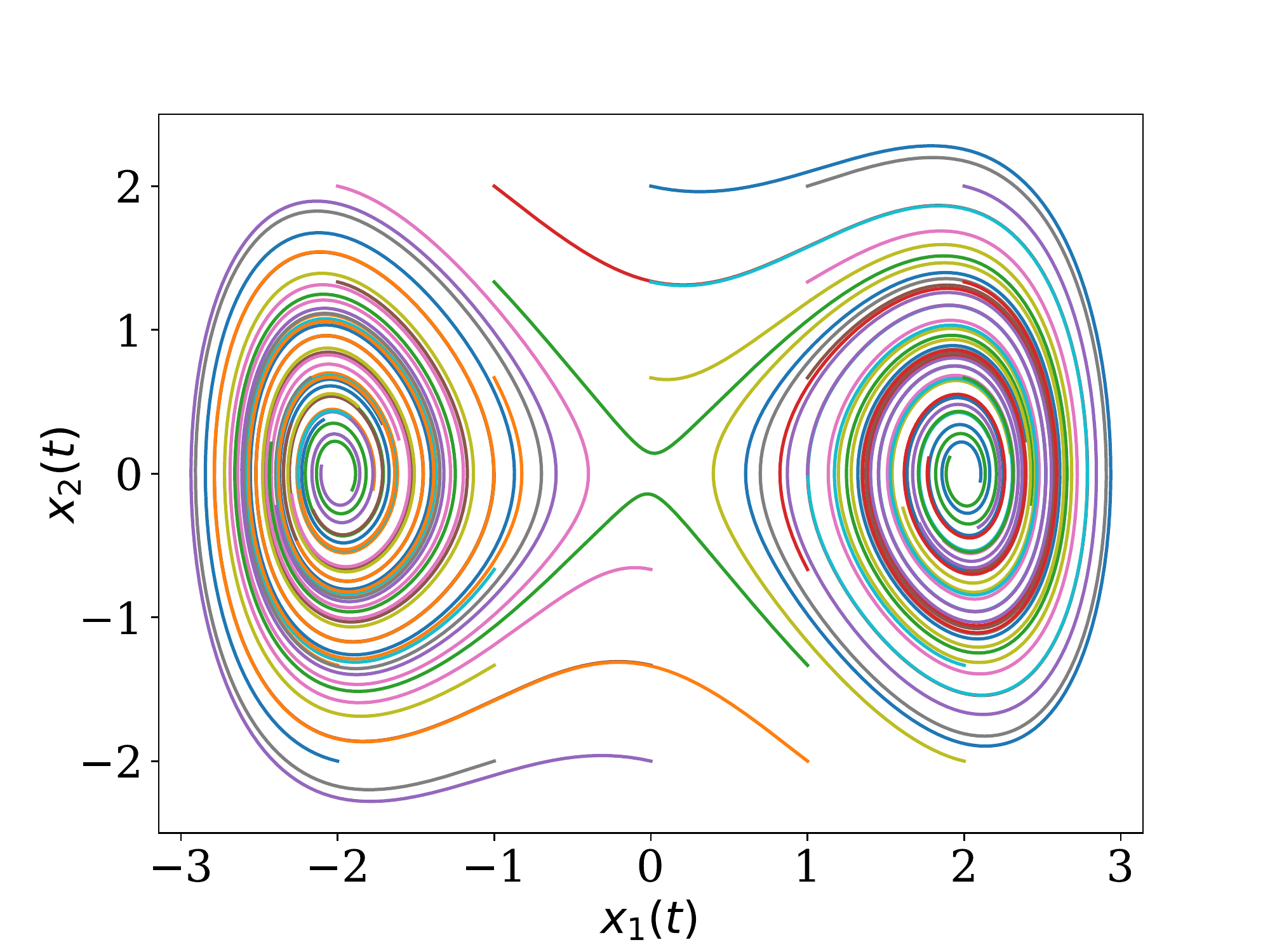} \hspace{-0.25cm}
\includegraphics[trim=40 0 50 0,clip,height=2.55in]{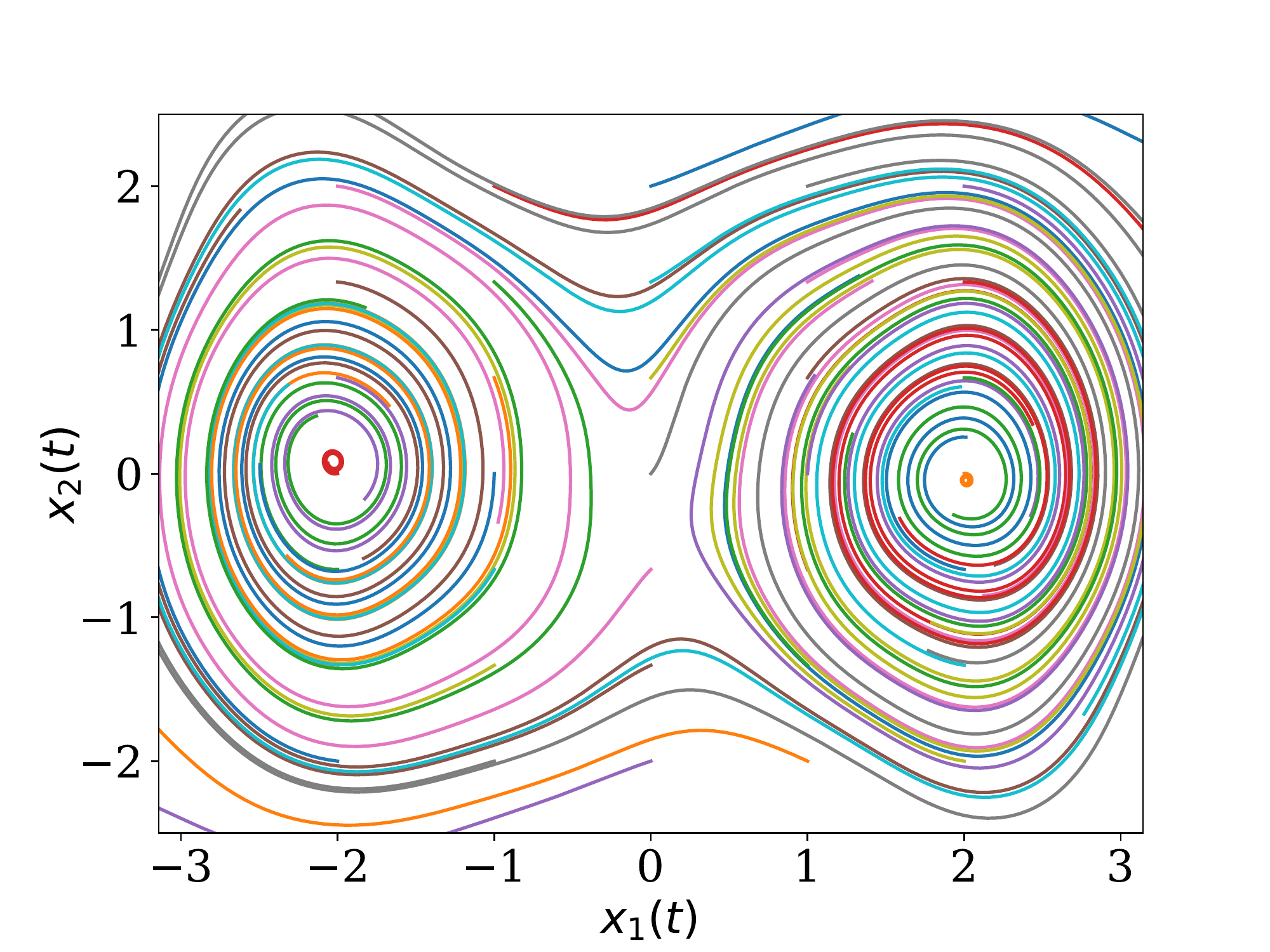}
\caption{\emph{Deterministic} phase portraits corresponding to the ground truth polynomial field $\widetilde{\bbf}(x) = (x_2, (2/5) x_1(4-x_1^2) - x_1/4)$ (left) and learned Fourier model $\bbf$ (right).}
\label{fig:polydblwell2d}
\end{figure}

\textbf{Two-Dimensional Trigonometric Vector Fields ($d=2$).} We continue with the difficult $\alpha=1$ case.  For the first two two-dimensional problems we consider, the training data consists of $n_T = 100$ trajectories of length $N = 101$ with $\Delta t = 0.4$.   We set $\nu = 4000$ so that our method's internal time step is $h = 10^{-4}$. The diffusion vector is set to $\mathbf{g} = (0.1, 0.1)$.

For data sets from two different ground truth $\widetilde{\bbf}$ fields (described below), we use our methods to estimate $\btheta$ with $J = 4$.  Note that because $d=2$, the complex array $\btheta$ is of size $(2J + 1) \times (2J +1) \times 2$.  In Figure \ref{fig:twodimresults}, we use $\theta_q$ as shorthand for $\btheta[:,:,q]$, the matrix of two-dimensional Fourier series coefficients for the $q$-th component of the vector field $\bbf$. To avoid overfitting and to promote sparsity of $\btheta$, we add an $\mathscr{L}^1$ regularization term $\mu \| \btheta \|_1$ to our MMD loss , together with a corresponding subgradient of this term in our gradient function.  We set $\mu = 1$ and do not take any steps to optimize/tune this hyperparameter. 

In Figure \ref{fig:twodimresults}, we plot the true and learned results for  $\widetilde{\bbf}(x_1, x_2) = (\sin x_2, -\sin x_1)$ (single well potential, left of the vertical bar) and  $\widetilde{\bbf}(x_1, x_2) = (\sin x_2, -V'(x_1))$ with $V(x) = ((\sin x/2)^2 - 4)^2/10$ (double well potential, right of the vertical bar).  Each plot is a heatmap of $|\theta_q|$.  Overall, we see close agreement between learned and true $\theta$ matrices in each case.  The MAE between learned and true Fourier coefficients is $6.4 \times 10^{-5}$ (left) and $6.2 \times 10^{-3}$ (right).  As the true $\theta_q$ matrices are highly sparse, we are confident that closer agreement is possible with our current setup, with a combination of more data and improved tuning of the hyperparameter $\mu$.

\textbf{Two-Dimensional Polynomial Vector Fields ($d=2$).} We now turn to two cases designed to challenge our current approach.  In both cases, the true $\widetilde{\bbf}$ is polynomial in nature, yet we seek a Fourier model $\bbf$. Continuing with $\alpha = 1$, we consider the polynomial double-well potential $V(x) = (x^2 - 4)^2/10$ and associated vector field with dissipation, $\widetilde{\bbf}(x_1, x_2) = (x_2, -V'(x_1)- x_1/4)$.

In this case, we began with a set of $n_T = 100$ trajectories each of length $N = 41$ with $\Delta t = 1.0$.   We set $\nu = 100$ so that our method's internal time step is $h = 10^{-2}$.  The diffusion vector is set to $\mathbf{g} = (0.1, 0.1)$.  Unlike the periodic two-dimensional vector fields above, this $\widetilde{\bbf}$ is unbounded, leading to trajectories with massive range in physical space.  We restricted attention to trajectories that stayed within the box $[-2 \pi, 2\pi]^2$, eliminating all but $29$ of the $100$ trajectories.

Because $n_T = 29$ is a small number of trajectories, we did not obtain reasonable results using the \emph{averaged} empirical characteristic function (\ref{eqn:empiricalcharfun}) as our target in the MMD loss.  Hence we replaced (\ref{eqn:empiricalcharfun}) with $\widetilde{\psi}^k(\bs, t_j) = \exp(i \bs^T \bX^k_j - (1/8) \bs^T \bs)$ for each $k = 1, \ldots, n_T$.  Here each of the $n_T$ trajectories is being treated as a target on its own, leading to one squared loss per trajectory.  Averaging the resulting per-trajectory squared losses resulted in only superficial modifications to the loss (\ref{eqn:discretemmdloss}) and associated adjoint method.    Note also the dash of Gaussian regularization added to each empirical characteristic function, designed to force $\widetilde{\psi}^k$ to decay to zero.  This improves performance of the characteristic function evolution method. 

We set $J = 8$ in our Fourier model $\bbf$ and train.  In Figure \ref{fig:polydblwell2d}, we plot \emph{deterministic} phase portraits for the ODE systems $\dot{\bx} = \widetilde{\bbf}(\bx)$ (left, ground truth) and $\dot{\bx} = \bbf(\bx)$  (right, learned).  We note that these vector fields are in qualitative agreement, with two stable fixed points located at approximately $(\pm 2, 0)$ separated by a saddle near the origin.  We believe that this qualitative agreement is sufficient for many purposes, e.g., using the identified system $\bbf$ for prediction and/or control.  These results were possible even with coarse temporal resolution in the training data ($N=41$ and $\Delta t=1.0$). 

Next, we consider the Maier-Stein vector field $\widetilde{\bbf}(\bx) = (x_1 - x_1^3 - x_1 x_2^2, -(1 + x_1^2) x_2)$.  We set $\alpha = 1.5$ and $\bg = (1, 1)$ to match \cite{fang2022end}.  We began with a set of $n_T = 100$ trajectories each of length $N = 11$ with $\Delta t = 1.0$. We set $\nu = 1000$ so that our method's internal time step is $h = 10^{-3}$. Restricting attention to trajectories confined to $[-2 \pi, 2\pi]^2$, we retained $n_T = 58$ trajectories.  As our target, we use the empirical characteristic functions $\widetilde{\psi}^k(\bs, t_j) = \exp(i \bs^T \bX^k_j)$ for $k = 1, \ldots, n_T$.  As $\alpha=1.5$ and $\bg = (1,1)$ in this case, we deemed it unnecessary to include a Gaussian regularization in $\widetilde{\psi}^k$.

\begin{figure}[t]
\centering
\includegraphics[trim=10 0 50 0,clip,width=2.1in]{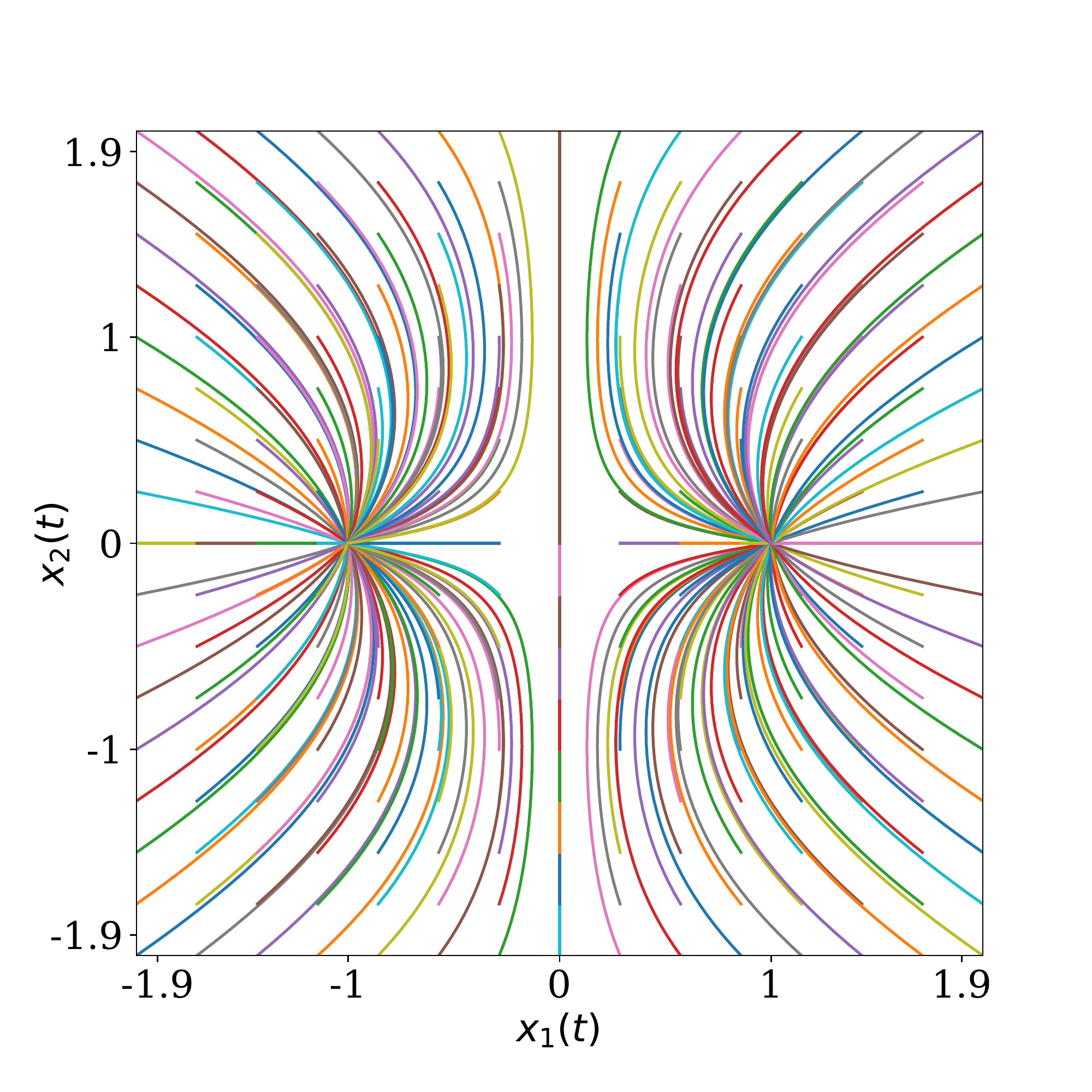} \hspace{-0.7cm}
\includegraphics[trim=10 0 50 0,clip,width=2.1in]{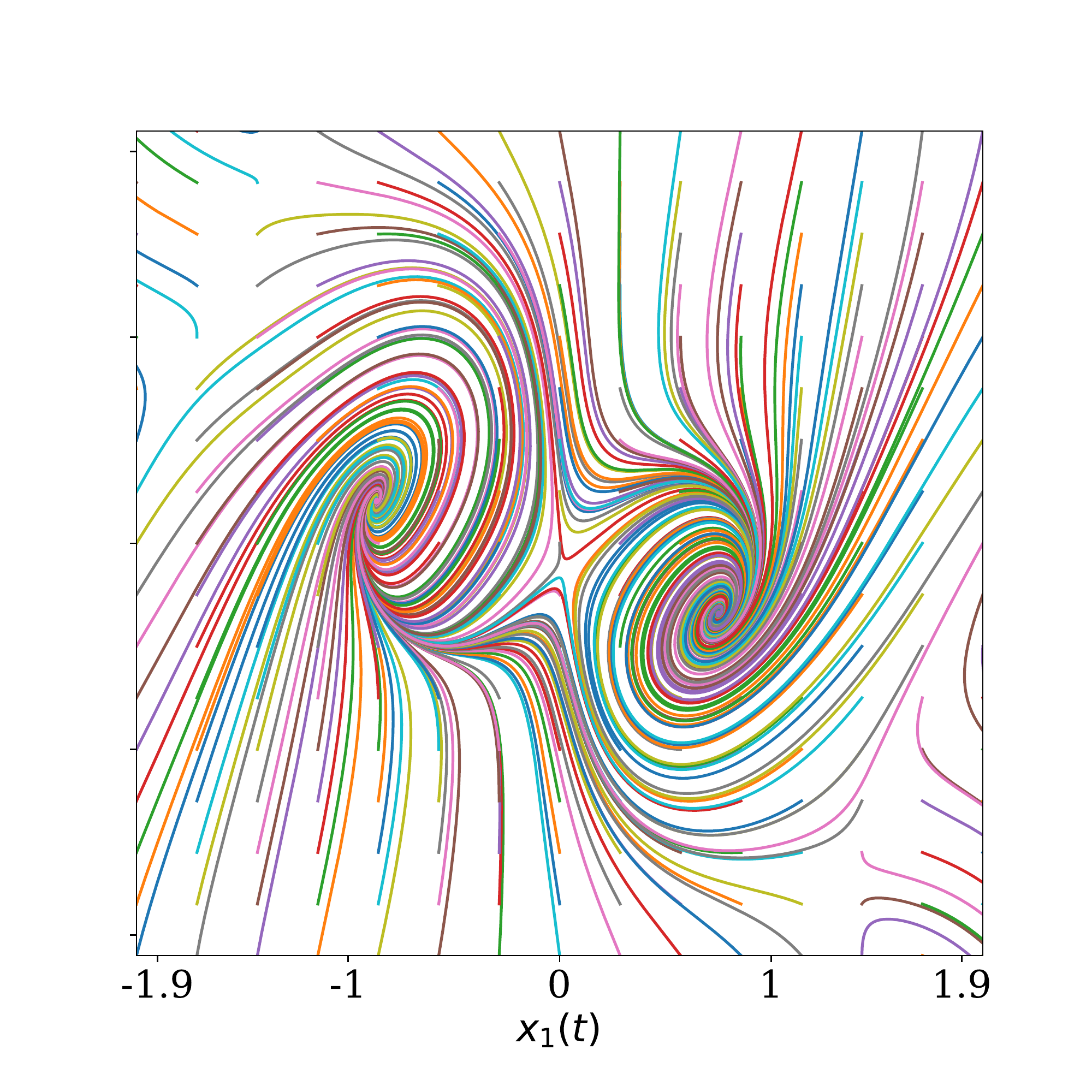} \hspace{-0.7cm}
\includegraphics[trim=10 0 50 0,clip,width=2.1in]{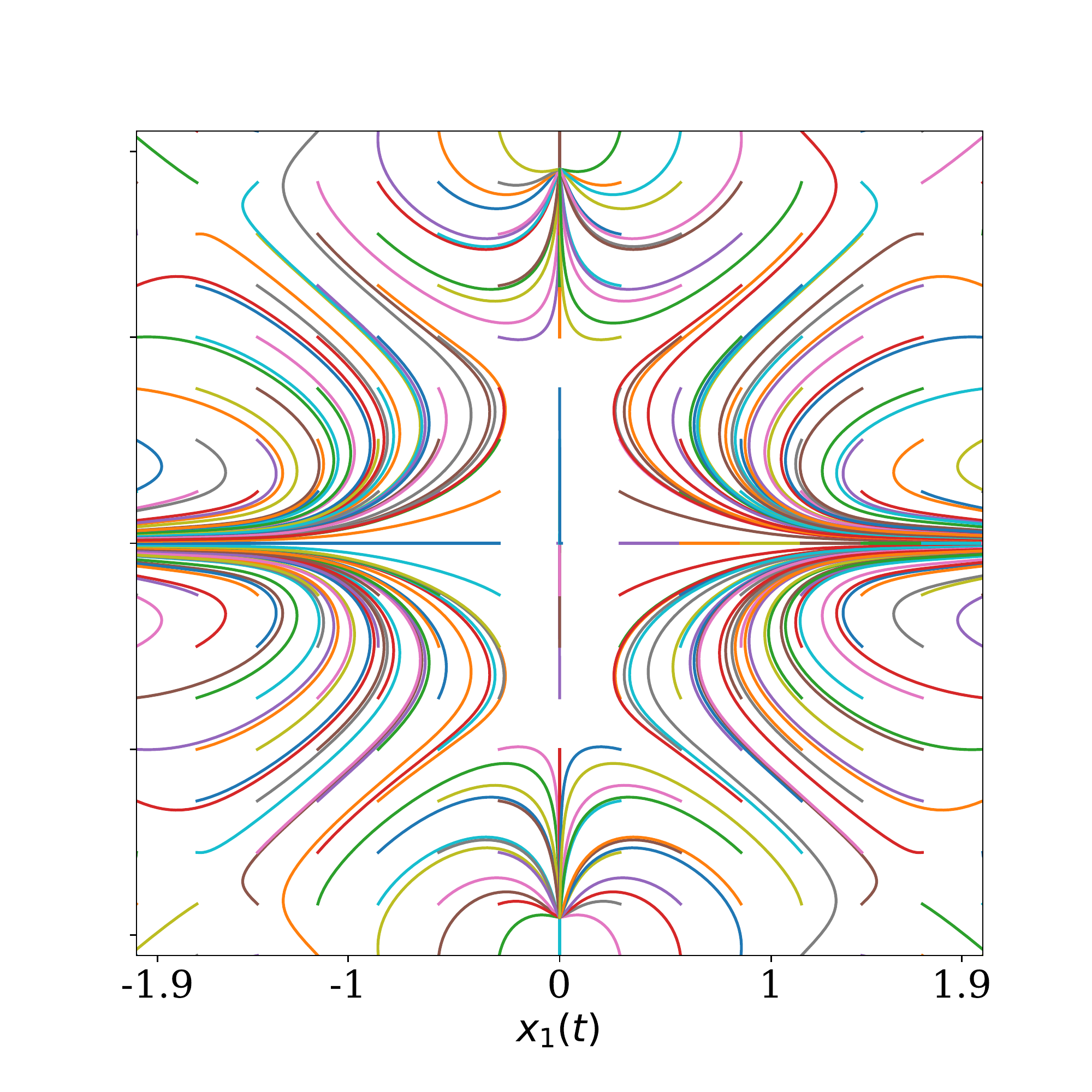} \caption{\emph{Deterministic} phase portraits illustrating the ground truth Maier-Stein vector field $\widetilde{\bbf}(\bx) = (x_1 - x_1^3 - x_1 x_2^2, -(1 + x_1^2) x_2)$ (left), and two learned vector fields $\bbf$, one with no symmetries enforced (middle) and one with even/odd symmetries that match those of  $\widetilde{\bbf}$ (right).}
\label{fig:MaierStein}
\end{figure}

We again set $J = 8$ in our Fourier model $\bbf$ and train.  In Figure \ref{fig:MaierStein}, we plot \emph{deterministic} phase portraits for the ODE systems $\dot{\bx} = \widetilde{\bbf}(\bx)$ (left, ground truth ) and $\dot{\bx} = \bbf(\bx)$  (middle, learned).  Note that even with the sparse-in-time training set, we learn a vector field with two stable fixed points located roughly near $(\pm 1, 0)$ as in $\widetilde{\bbf}$.  Based on this experiment, we realized that $\widetilde{\bbf}$ enjoys symmetries that stem from the fact that $\widetilde{f}_1$ is odd in $x_1$ and even in $x_2$, while $\widetilde{f}_2$ is even in $x_1$ and odd in $x_2$.  We modified the Fourier representation (\ref{eqn:L2drifthat}) to incorporate these symmetries.  As before, this required only superficial changes to the loss (\ref{eqn:discretemmdloss}) and associated adjoint method.  Retraining, we obtain a vector field $\bbf$ associated with the phase portrait on the far right of Figure \ref{fig:MaierStein}.  We now see improved agreement with the left panel in terms of the vertical separatrix at $x=0$ together with the shapes of the integral curves flowing into the $y=0$ axis.

Again, these results are possible even with coarse temporal resolution in the training data ($N=11$ and $\Delta t=1.0$).  We conjecture that a more scalable implementation of our algorithm in the $d=2$ case, enabling us to increase the volume of training data and increase spatial discretization parameters such as $L$, $M$, and $J$, will lead to improved learning of polynomial vector fields.

%%%%%%%%%%%%%%%%%%%%%%%%%%%%%%%%%%%%%%%%%%%%%%%%%%%%%%%%%%%%%%%%%%%%%%%%%%%%%%%%
\section{Conclusion}
The characteristic function evolution method enables us to accurately and stably evolve L\'{e}vy $\alpha$-stable dynamical systems forward in time.  When coupled with the MMD loss function and the adjoint method, we have an end-to-end method for system identification.  In this paper, we derived the method for the first time and demonstrated its potential.  When the ground truth field is periodic, we succeed in identifying it using our Fourier representation, even for two-dimensional systems in the challenging $\alpha = 1$ (Cauchy noise) case.  Learned vector fields can capture key qualitative features of ground truth polynomial vector fields. In future work, we plan to expand our method to incorporate other types of models for $\bbf$, e.g., polynomial and neural network models.  It seems likely that by combining Fourier space ideas proposed in the present paper with modern techniques from the literature, one may derive still superior system identification methods.  We hope the present work enables further research in this direction.

%%%%%%%%%%%%%%%%%%%%%%%%%%%%%%%%%%%%%%%%%%%%%%%%%%%%%%%%%%%%%%%%%%%%%%%%%%%%%%%%
\clearpage

\acks{This research was partially supported by NSF DMS-1723272, and also benefited from computational resources that include the Pinnacles cluster at UC Merced (supported by NSF OAC-2019144) and  Nautilus, supported by the Pacific Research Platform (NSF ACI-1541349), CHASE-CI (NSF CNS-1730158), Towards a National Research Platform (NSF OAC-1826967), and the University of California Office of the President. The author expresses gratitude for discussions with Arnold D. Kim on an earlier version of the project, and conversations with Hua Hsu that motivated completion of the work.}

\bibliography{main}

\clearpage

\section*{Appendix}

\subsection{Extended Literature Review}
\label{sect:litreview}
As mentioned in Section \ref{sect:intro}, Fourier duality gives us the intuition that characteristic function approaches to (\ref{eqn:sde}) may be both mathematically natural and computationally effective.  We find validation of these ideas in recent work that addresses problems for  (\ref{eqn:sde}) that are adjacent to but yet different from drift identification.  For instance, \cite{Talebi2023} combined fractional-order calculus with a characteristic function framework to solve filtering problems for (\ref{eqn:sde}).  One may be able to combine these ideas with those proposed here to pursue drift identification from noisy observations.
 
 Our characteristic function evolution method is an alternative to the numerical solution of the fractional/non-local Fokker-Planck-Kolmogorov equation associated with (\ref{eqn:sde}).  Suppose we take Fourier transforms of both sides of such an equation.  We would then derive the differential version of the integral method that we pursue; in the literature, this differential equation is called the \emph{spectral counterpart} to the Fokker-Planck-Kolmogorov or Einstein-Smoluchowski equation \citep{Cottone_2011,Alotta2014,ALOTTA2015265,DiMatteo2017,DiPaola2020,Yang2021}.

In much of this literature, time-dependent characteristic functions are used as intermediaries either to compute time-dependent densities (via inverse Fourier transform), or to establish connections between different types of evolution equations.  Only \cite{Cottone_2011} derives a method that, as in our method, evolves the time-dependent characteristic function forward in time without any computations in physical space.  Still, in the above literature, the full form of the drift and diffusion fields is assumed to be known \emph{a priori} and the methods are used entirely for forward simulation.

In order for Fourier space methods to succeed in system identification, they should be able to compute the time-dependent characteristic function stably and accurately \emph{even when the drift is approximated badly}, as may occur during the initial steps of an optimization loop.  Our derivation, which begins with the Chapman-Kolmogorov equation in integral form (\ref{dtq}) and ends with a closed-form integral equation (\ref{eqn:charfunevol}) with no temporal or spatial derivatives, is designed to produce such a method. 

\subsection{Illustrative/Theoretical Properties}
\label{sect:properties}

\subsubsection{Global Bound}
\label{sect:globalbound}
Using $|e^{i \bs^T \bx}| = 1$
and $p(\bx) \geq 0$, we have
\[
| \psi(\bs) | = \left| \int_{\bx \in \bbR^d} e^{i \bs^T \bx} p(\bx) \, \mathrm{d}\bx. \right| \leq \int_{\bx \in \bbR^d} p(\bx) \, \mathrm{d}\bx = 1.
\]
Because $\psi(\mathbf{0}) = 1$, we see that $\|\psi \|_\infty = 1$.  This global bound supports the strategy of using characteristic functions in a numerical scheme. 

\subsubsection{Concrete Example of Characteristic Function Evolution}
\label{sect:concrete}
Let us focus attention on the one-dimensional ($d=1$) case.  For one particular choice of $f$ and $G$, we can use (\ref{eqn:ctq}) to solve for the time-dependent characteristic function of the solution of (\ref{eqn:sde}).  We present this example to build intuition regarding (\ref{eqn:ctq}).  In what follows, we use 
\[
\frac{1}{(2\pi)^d} \int_{\by \in \bbR^d} e^{i (\bs - \bu)^T y} \, \mathrm{d}\by = \delta(\bs - \bu),
\]
which was also used to compute the $O(h^0)$ term in (\ref{eqn:ourke}).  Returning to $\widetilde{K}$ defined by (\ref{eqn:kdef}), we see that
\[
\lim_{h \to 0}
\widetilde{K}(s,u) = \delta(s - u).
\]
Now let us briefly switch gears.  The following ordinary differential equation (ODE) is one of the easiest to solve:
\[
\dot{x} = -x.
\]
Given $x(0) = x_0$, the solution is $x(t) = x_0 e^{-t}$.  The ODE has a globally stable, attracting fixed point at $x = 0$.  This ODE is in fact a special, noiseless case of our L\'{e}vy SDE, with $f(x) = -x$ and $G(x) \equiv 0$.  The simplest way to reintroduce noise is to take $G(x) = g > 0$, a constant.  In this case, the kernel becomes
\begin{align}
\widetilde{K}(s,u) &= \frac{1}{2\pi} \exp{\left(  -h g^\alpha |s|^{\alpha} \right)} \int_{y=-\infty}^{\infty}e^{is y \left(1 - h\right)} e^{-iuy}\, dy \nonumber \\
\label{eqn:OUprop}
 &= \exp{\left(  -h g^\alpha |s|^{\alpha} \right)} \delta( s(1-h) - u).
\end{align}
With this kernel the evolution equation (\ref{eqn:ctq}) becomes
\[
\psi_{n+1}(s) = \exp{\left(  -h g^\alpha |s|^{\alpha} \right)} \psi_n( s(1-h) ).
\]
These relationships telescope, starting at $\psi_n$ and going back to the initial condition $\psi_0$:
\begin{align*}
\psi_{n}(s) &= \exp{\left(  -h g^\alpha |s|^{\alpha} \right)} \psi_{n-1}( s(1-h) ) \\
\psi_{n-1}(s) &= \exp{\left(  -h g^\alpha |s|^{\alpha} \right)} \psi_{n-2}( s(1-h) ) \\
&\vdots \\
\psi_{2}(s) &= \exp{\left(  -h g^\alpha |s|^{\alpha} \right)} \psi_1( s(1-h) ) \\
\psi_{1}(s) &= \exp{\left(  -h g^\alpha |s|^{\alpha} \right)} \psi_0( s(1-h) ).
\end{align*}
Putting things together, we obtain
\[
\psi_{n}(s) = \exp \left( -h g^\alpha |s|^\alpha \sum_{j=0}^{n-1} |1-h|^{j \alpha} \right) \psi_0( s (1-h)^n ).
\]
Let $n h = t$ for some time $t > 0$.  
Fixing $t$ and taking $h \to 0$, we obtain
\[
\psi(s, t) = \exp{ \left( - g^\alpha |s|^{\alpha} \alpha^{-1} (1 - e^{-t \alpha}) \right)} \psi_0( e^{-t} s ).
\]
When $\alpha = 2$, this is the Fourier transform of the Ornstein-Uhlenbeck probability density function.  When $\alpha=2$, the SDE with drift $f(x) = -x$ and constant $G$ is indeed the Ornstein-Uhlenbeck SDE driven by standard Brownian motion.  The upshot: we have used (\ref{eqn:ctq}) to solve this SDE for all $\alpha$, not only $\alpha = 2$.

\subsubsection{Concrete Example of Kernel Expansion}
\label{sect:kernexpanexample}
Continuing with the one-dimensional case, let us now consider $f(x) = \sin x$ and $G(x) \equiv g > 0$.  Using
\begin{equation}
\label{eqn:sinhat}
\frac{1}{2 \pi} \widehat{f}(k) = \frac{1}{2 \pi} \int_{y=-\infty}^\infty e^{i k y} \sin y \, dy = \frac{i}{2} \left[ \delta(k-1) - \delta(k+1) \right].
\end{equation}
and
\begin{equation}
\label{eqn:sinsqhat}
\frac{1}{2 \pi} \widehat{f^2}(k) = \frac{1}{2 \pi} \int_{y=-\infty}^\infty e^{i k y} \sin^2 y \, dy = \frac{1}{4} \left[ -\delta(k-2) + 2 \delta(k) - \delta(k+2) \right],
\end{equation}
we can compute the kernel expansion  (\ref{eqn:ourke}) up to second-order in $h$:
\begin{multline}
\label{eqn:kesin}
\widetilde{K}(s,u) = \exp(-h |s g|^{\alpha})  \biggl[ \delta(s-u) - \frac{1}{2} s h \bigl( \delta(s-u-1) - \delta(s-u+1) \bigr) \\ - \frac{1}{8} s^2 h^2 \bigl( -\delta(s-u-2) + 2 \delta(s-u) - \delta(s-u+2) \bigr) \biggr]
\end{multline}
Then using this kernel expansion in (\ref{eqn:ctq}), we obtain
\begin{multline}
\label{eqn:ctqsin}
\psi_{n+1}(s) = \exp(-h |s g|^{\alpha}) \biggl[
\Bigl( 1 - \frac{1}{4} s^2 h^2 \Bigr) \psi_n(s) \\
- \frac{1}{2} s h \Bigl( \psi_n(s-1) - \psi_n(s + 1) \Bigr)
+ \frac{1}{8} s^2 h^2 \Bigl( \psi_n(s-2) + \psi_n(s + 2) \Bigr) \biggr]
\end{multline}
 We can apply methods such as collocation to use (\ref{eqn:ctqsin}) to evolve the characteristic function $\psi_n$ forward in time starting from $\psi_0$.  
More generally, \emph{if we know} the full form of an SDE, including the drift and diffusion functions, we may be able to derive from (\ref{eqn:ctq}) a customized numerical method for that particular SDE.  While we do not pursue this in the present paper, we see scope for further numerical analysis of this approach as a method to solve L\'{e}vy-driven SDE. 

Note that the coefficients that appear on the right-hand side of (\ref{eqn:ctqsin}) are in fact Taylor expansions of Bessel functions of the first kind.  Consider the exact kernel (\ref{eqn:kdef}) with $f(y) = \sin y$ and apply the Jacobi-Anger expansion to obtain:
\begin{align}
\widetilde{K}(s,u) &= \frac{1}{2 \pi}  \exp(-h |s g|^{\alpha})  \int_{y=-\infty}^\infty e^{i(s-u)y} e^{i s h \sin y } \, dy \nonumber \\
 &= \frac{1}{2 \pi}  \exp(-h |s g|^{\alpha})  \int_{y=-\infty}^\infty e^{i(s-u)y} \sum_{n=-\infty}^\infty J_n(s h) e^{i n  y} \, dy \nonumber \\
 \label{eqn:exactsinkernel}
 &=  \exp(-h |s g|^{\alpha}) \sum_{n=-\infty}^\infty J_n(s h) \delta(s - u + n) 
\end{align}
Now note that
\begin{align*}
J_0 (sh) &= 1 - \frac{1}{4} s^2 h^2 + O(h^4) \\
J_{\pm 1} (sh) &= \pm \frac{1}{2} s h + O(h^3) \\
J_{\pm 2} (sh) &= \frac{1}{8} s^2 h^2 + O(h^4)
\end{align*}
For $|n| \geq 3$, the expansion of $J_{n}(s h)$ begins with a term that is at least cubic in $s h$, and hence can be ignored for our purposes.  Now substituting these Bessel function expansions into (\ref{eqn:exactsinkernel}) and ignoring terms for which $|n| \geq 3$, we obtain precisely the same result as the kernel expansion (\ref{eqn:kesin}).

\subsection{On the Accuracy and Stability of the Fully Discrete Scheme (\ref{eqn:psiupdate2})}
\label{sect:accstab}
To derive (\ref{eqn:psiupdate2}), we expanded in the time step $h$ up to second order.  Numerical analysis of the Euler-Maruyama method for SDE driven by L\'{e}vy $\alpha$-stable processes has established weak convergence at rates no better than $O(h)$ \citep{Jacod2004, MIKULEVICIUS20111720}.  Our intuition is that because our approximation has $O(h^3)$ truncation error, even after $T/h$ time steps, the error of our approximation will be dominated by the error of the Euler-Maruyama method itself.  As weak convergence is related to convergence of densities and characteristic functions, we expect to establish in future work that (\ref{eqn:charfunevol}) converges at the same weak convergence rate of Euler-Maruyama.

\begin{figure}[t]
\begin{center}
\includegraphics[width=3.25in]{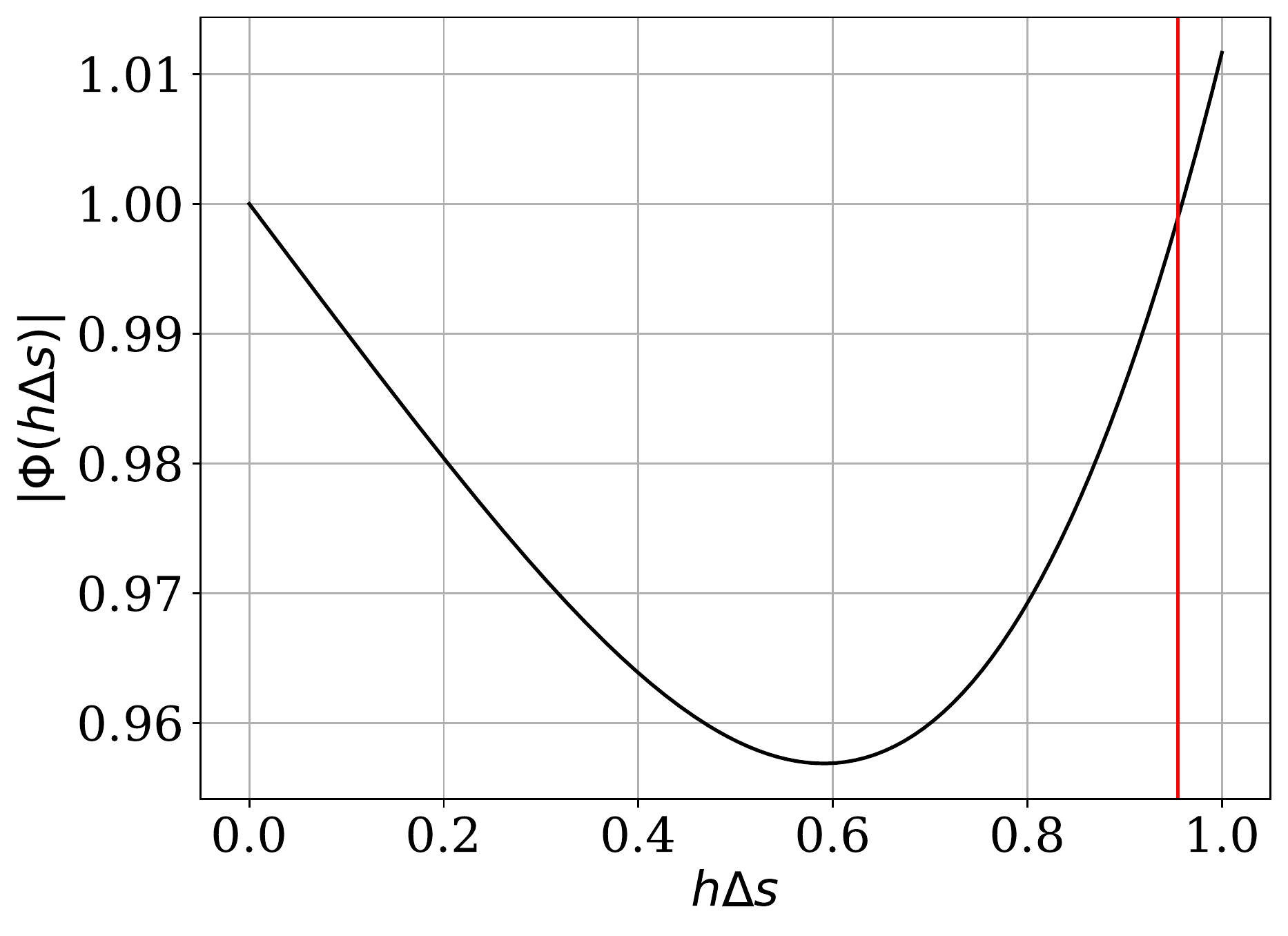}
\end{center}
\caption{For $g = 1/10$, we plot $|\Phi(h \Delta s)|$ as a function of $h \Delta s$ and find that $|\Phi(h \Delta s)| < 1$ for $0 < h \Delta s < (h \Delta s)^\ast$ with $(h \Delta s)^\ast \approx 0.955$ drawn in red.  This gives a rough estimate of the stability region for (\ref{eqn:psiupdate2}).}
\label{fig:stability}
\end{figure}

We also see from (\ref{eqn:psiupdate2}) that the time step $h$ always occurs together with the grid spacing $\Delta s$.  The quantity $h \Delta s$ can therefore be expected to play a role in accuracy and stability.  One way to obtain a rough guide for stability is to analyze (\ref{eqn:psiupdate2}) in the scenario where $\btheta$ is a Kronecker delta, $\mathbf{j}$ is a unit vector, and $\psi_k$ is constant in space.  In this case, (\ref{eqn:psiupdate2}) reduces to $z_{k+1} = \Phi(h \Delta s) z_k$ with
\[
\Phi(w) = e^{-w g^\alpha} (1 + i w - w^2/2).
\]
Stability is assured for all $w = h \Delta s$ such that $|\Phi(w)| \leq 1$.  As one might expect, as the diffusion constant $g$ increases, stability is guaranteed for larger values of $w$.  For instance, when $g=1$, $|\Phi(w)| \leq 1$ for all $w \geq 0$, with equality only at $w=0$.  For a more realistic value, such as $g=1/10$, we find that $|\Phi(w)| < 1$ for $0 < w < 0.955$, as depicted in Figure \ref{fig:stability}.  At least in this idealized setting, stability is possible at reasonable values of $h \Delta s$, even though  (\ref{eqn:psiupdate2}) is a fully explicit scheme.

\subsection{Adjoint Method Details}
\label{sect:adjdeets}
Here we detail an adjoint method designed to minimize the MMD loss $\Lambda(\btheta)$ (\ref{eqn:discretemmdloss}) subject to the dynamics (\ref{eqn:psiupdate2}).

To simplify the notation, let $\bpsi_t \in \mathbb{C}^{(2M+1)^d}$ denote the complex vector that collects all entries of $\psi(\bj \Delta s, t; \btheta)$.  Then the evolution equation (\ref{eqn:psiupdate2}) can be written in the following abstract form, with $k = 0, \ldots, \nu-1$:
\begin{equation}
\label{eqn:dynamics}
\bpsi_{t_j + (k+1) h} = \mathscr{P}(\btheta) \bpsi_{t_j + k h}
\end{equation}
Here $\mathscr{P}$ is our discrete-space, discrete-time propagator.  One can obtain the $(\bj, \bj')$-th entry of $\mathscr{P}$ by differentiating the right-hand side of (\ref{eqn:psiupdate2}) with respect to $\psi_k(\bj')$.

With this, we can formulate the Lagrangian as
\begin{multline}
\label{eqn:lagrangian}
\mathcal{L}(\bpsi, \blambda, \btheta) = \frac{1}{2} \sum_{j=1}^{N} \Bigl\| \bpsi_{j} - \widetilde{\bpsi}_j \Bigr\|^2 - \Re \sum_{j=0}^{N-1} \sum_{k = 0}^{\nu - 1} \blambda_{t_j + (k+1) h}^\dagger \left(\bpsi_{t_j + (k+1) h} - \mathscr{P}(\btheta) \bpsi_{t_j + k h} \right)
\end{multline}
Here $\blambda_{t_j + (k+1)h} \in \mathbb{C}^{(2M+1)^d}$ is a Lagrange multiplier that enforces the dynamical constraint (\ref{eqn:dynamics}).  Taking variations with respect to $\bpsi_j$, we obtain
\begin{align*}
\delta \mathcal{L} &= \Re \sum_{j=1}^N (\bpsi_j - \widetilde{\bpsi}_j )^\dagger \delta \bpsi_{t_j} - \Re \sum_{j=0}^{N-1} \sum_{k = 0}^{\nu - 1} \blambda_{t_j + (k+1) h}^\dagger \left(\delta \bpsi_{t_j + (k+1) h} - \mathscr{P}(\btheta) \delta \bpsi_{t_j + k h} \right) \\
&= \Re \sum_{j=1}^N (\bpsi_j - \widetilde{\bpsi}_j )^\dagger \delta \bpsi_j  - \Re \sum_{j=0}^{N-1} \sum_{k=1}^{\nu} \blambda^\dagger_{t_j + k h} \delta \bpsi_{t_j + k h} + \Re \sum_{j=0}^{N-1} \sum_{k = 0}^{\nu - 1} \blambda_{t_j + (k+1) h}^\dagger \mathscr{P}(\btheta) \delta \bpsi_{t_j + k h}
\end{align*}
For optimality, we want $\delta \mathcal{L}$ to vanish for all variations $\delta \bpsi$.  Note that the $k = \nu$ terms are present only in the second sum on the right-hand side, not in the third sum.  At $k = \nu$, since $\nu h = \Delta t$, we have $t_j + \nu h = t_{j+1}$.  Therefore, at $k = \nu$, $\delta \bpsi_{t_j + k h} = \delta \bpsi_{t_{j+1}}$.  These variations vanish if we set
\begin{equation}
\label{eqn:fincond}
\blambda_{t_j} = \bpsi_j - \widetilde{\bpsi}_j,
\end{equation}
for $j = 1, \ldots, N$.  The remaining variations vanish if we set
\begin{equation}
\label{eqn:adjoint}
\blambda^\dagger_{t_j + k h} = \blambda_{t_j + (k+1) h}^\dagger \mathscr{P}(\btheta).
\end{equation}
We recognize (\ref{eqn:fincond}) as the final condition for (\ref{eqn:adjoint}), the backward-in-time adjoint equation corresponding to the forward dynamics (\ref{eqn:dynamics}).

We can now outline a procedure to compute $\nabla_{\btheta} \Lambda$, the gradient of the loss function with respect to the model parameters.  Given a trial value of $\btheta$, we solve (\ref{eqn:dynamics}) forward in time using the empirical characteristic functions at times $t_j$ (for $j = 0, \ldots, N-1)$ as initial conditions---see (\ref{eqn:empiricalcharfun}) and surrounding discussion.  We then solve (\ref{eqn:adjoint}) backward in time with final condition (\ref{eqn:fincond}).  This enables us to evaluate
\begin{equation}
\label{eqn:parametergrad}
\nabla_{\btheta} \mathcal{L} =  \Re \sum_{j=0}^{N-1} \sum_{k = 0}^{\nu - 1} \blambda_{t_j + (k+1) h}^\dagger  \nabla_{\btheta} \mathscr{P}(\btheta) \bpsi_{t_j + k h}.
\end{equation}
That this gradient equals $\nabla_{\btheta} \Lambda$ is a consequence of the Lagrange multiplier theorem.  To compute the right-hand side (\ref{eqn:parametergrad}), it suffices to have a method to multiply the gradient $\nabla_{\btheta} \mathscr{P}(\btheta)$ by a vector $\bpsi$.  We obtain such a method by differentiating the right-hand side of (\ref{eqn:psiupdate2}) with respect to $\btheta$.  This yields
\begin{equation}
\label{eqn:gradPaction}
\frac{\partial}{\partial \theta^{\mathbf{r}}_q} \left[ \mathscr{P}(\btheta) \bpsi_{t} \right]_{\bj} = e^{-h \Delta s |\bj^T \bg|^\alpha} \biggl[ i (h \Delta s) j_q \psi_t(\bj + \mathbf{r} n_L) 
- (h \Delta s)^2 \sum_{\bk \in \mathcal{K}} j_q (\btheta^{\bk - \mathbf{r}}
)^T \mathbf{j} \, \psi_t(\bj + \bk n_L) \biggr]
\end{equation}

\subsection{Implementation Details}
\label{sect:implementationdetails}
For all tests, we use our own Python implementation of the characteristic function evolution method and associated adjoint method.  We have implemented the methods using Numba \citep{numba}, making use of both JIT, just-in-time compilation, and CUDA, enabling us to run our code on GPUs.  All Euler-Maruyama runs (to generate training data) were carried out using Mathematica.  We are in the process of making source code available at the URL \url{https://github.com/hbhat4000/levyL4DC}.

When we optimize in one-dimensional problems, we set the optimization tolerances to $10^{-9}$.  For two-dimensional problems, we set the tolerances to $10^{-3}$.

\paragraph{One-Dimensional Vector Fields.} To generate trajectories, we applied Euler-Maruyama with a time step of $10^{-3}$ for $4001$ steps, but only saved the solution every $100$ steps. 

When we apply the characteristic function evolution method (\ref{eqn:psiupdate2}), we use the following parameters: $L = 2$, $M = 1028$, and $n_L = 8$.  This implies that we track characteristic functions at $2M+1=2049$ points in Fourier space, with $\Delta s = 0.0625$.

\paragraph{Two-Dimensional Trigonometric Vector Fields.} To generate trajectories corresponding to the trigonometric vector fields with results plotted in Figure \ref{fig:twodimresults}, we applied Euler-Maruyama with time step $10^{-4}$ for $400001$ steps, saving the solution every $4000$ steps.

For each trajectory, the initial condition is $\bX_0 = (X_{0,0},0)$ where $X_{0,0}$ is drawn from a normal distribution with mean $0$ and standard deviation $1/3$.

To conserve memory, we track the characteristic function on a coarse grid with only $2M + 1 = 65$ points in Fourier space per dimension.  We set $L = 2$ and $n_L = 4$ so that $\Delta s = 0.125$. 

\paragraph{Two-Dimensional Polynomial Double Well Vector Field.} Initial conditions $\bX_0$ were chosen as described above for two-dimensional trigonometric vector fields. The Euler-Maruyama runs were also conducted as above, except that the solution was saved every $10^4$ steps.

We again track the characteristic function on a coarse grid with only $2M + 1 = 65$ points in Fourier space per dimension.  We set $L = 2$ and $n_L = 4$ so that $\Delta s = 0.125$. 

Note that we retain the $\mathscr{L}^1$ regularization term  $\mu \| \btheta \|_1$ with $\mu = 1$.  We seek to promote sparsity for this polynomial vector field because when we examine $\widetilde{\bbf}(\bx)$, we see that $\widetilde{f}_1$ depends only on $x_2$ and $\widetilde{f}_2$ depends only on $x_1$.  Hence the true $\widetilde{\btheta}$ coefficients will be sparse.

\paragraph{Maier-Stein Polynomial Vector Field.} Initial conditions $\bX_0$ were chosen from an equispaced $10 \times 10$ grid on the square $[-1, 1]^2$.  The Euler-Maruyama runs used a time step of $10^{-4}$ for $100001$ steps.  The solution was saved every $10000$ steps.

We again track the characteristic function on a coarse grid with only $2M + 1 = 65$ points in Fourier space per dimension.  We set $L = 2$ and $n_L = 8$ so that $\Delta s = 0.0625$. 

We keep the $\mathscr{L}^1$ regularization term  $\mu \| \btheta \|_1$ but relax $\mu$ to $0.1$. 

\subsection{Analyzing the Error for the One-Dimensional Polynomial Double Well}
\label{sect:onedimdblwell}
Here we continue the discussion of the right-most plot in Figure \ref{fig:onedimresults}, corresponding to the true (black) $\widetilde{f}(x) = x - x^3$ and the estimated (red) $f$.

For $|x| > 2$ (not plotted), the two curves diverge; the mean absolute error between estimated $\btheta$ and true $\widetilde{\btheta}$ coefficients is large.  We hypothesize that this occurs because (i) we are trying to model a polynomial vector field $\widetilde{f}$ using a Fourier series $f$, (ii) $J = 16$ Fourier modes is insufficient to capture the behavior of this polynomial $\widetilde{f}$, and (iii) the training data does not adequately explore phase space for $|x| > 2$.

\begin{figure}[t]
\begin{center}
\includegraphics[width=3in]{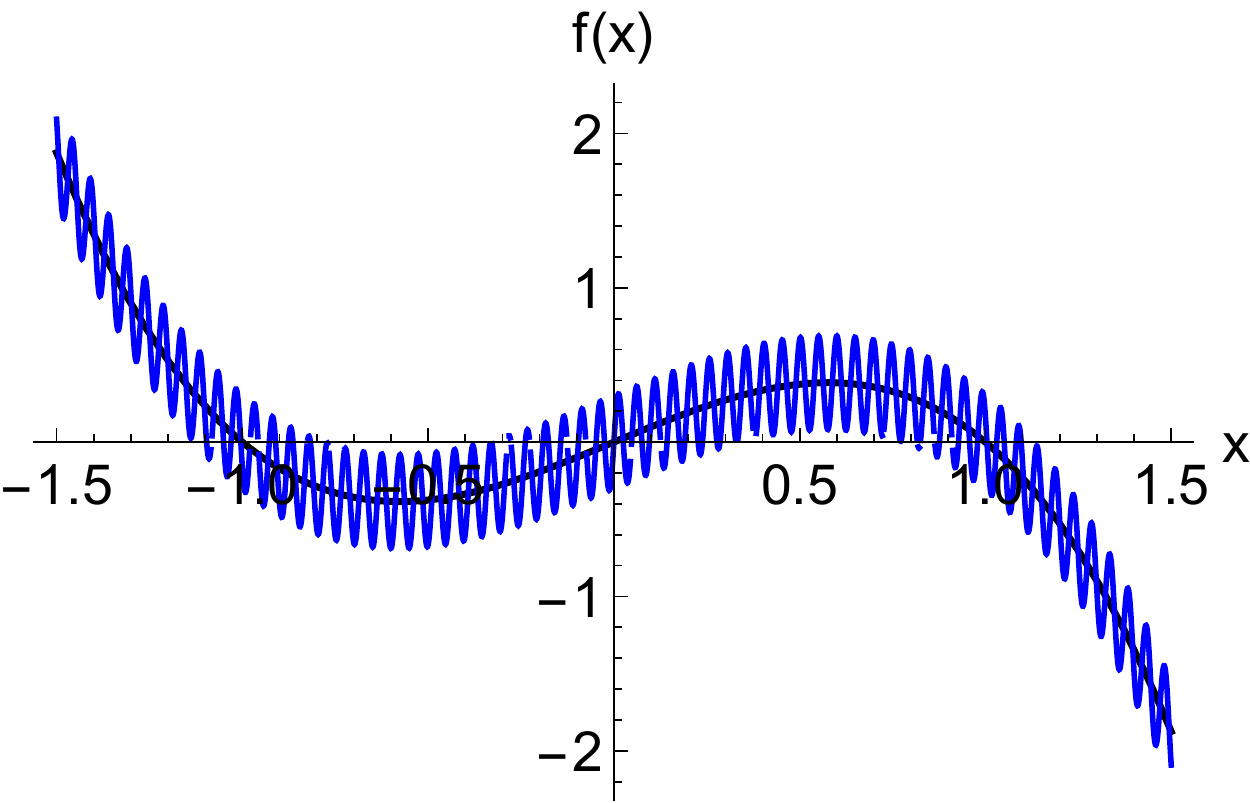}
\end{center}
\caption{Ground truth $\widetilde{f}(x) = x - x^3$ (in black) together with Fourier series approximation computed using exact Fourier coefficients (\ref{eqn:gtdblwell1d}) with $J=256$ modes (in blue).}
\label{fig:exactfourier}
\end{figure}

Let us focus on reason (i) above.  The Fourier coefficients $\widetilde{\btheta}$ corresponding to the ground truth field $\widetilde{f}$ satisfy $\widetilde{\theta}_0 = 0$ and for $j \neq 0$,
\begin{equation}
\label{eqn:gtdblwell1d}
\widetilde{\theta}_{j} = -\frac{2 i (-1)^j \left(\left(4 \pi ^2-1\right) j^2-24\right)}{j^3}
\end{equation}
It is clear that these coefficients decay to zero slowly, e.g., $|\widetilde{\theta}_j| \sim O(1/j)$ for $|j| \gg 1$.  Suppose we use these \emph{exact} Fourier coefficients in (\ref{eqn:L2drifthat}) with $J = 256$.  In Figure \ref{fig:exactfourier}, we have plotted the resulting Fourier series approximation (in blue) together with the ground truth $\widetilde{f}$ (in black).  Note the presence of oscillations that lead to an overall \emph{worse} approximation of $\widetilde{f}$ than we saw in the right-most panel of Figure \ref{fig:onedimresults}.  In short, the result we obtained in the right-most panel of Figure \ref{fig:onedimresults} with $J=16$ modes should be viewed as quite good, especially considering that $\alpha=1$ in the SDE (\ref{eqn:sde}).

With this in mind, we introduce another way to measure error, akin to pointwise-in-time test set error.  Using the true $\widetilde{f}$, we generate a set of trajectories $\mathcal{\widetilde{T}}$ starting from $X_0 = 0$.  For this same initial condition, we then recompute the trajectories with (i) the true $\widetilde{f}$, resulting in $\mathcal{T}'$, and (ii) the estimated $\widetilde{f}$, resulting in $\mathcal{T}$.  We then compute the median of median absolute errors (MMAE) and median of interquartile ranges (MIQR) between $\mathcal{\widetilde{T}}$ and $\mathcal{T}$, a natural measure of how well our learned $f$ predicts trajectories.  We compare this against the MMAE and MIQR between $\mathcal{\widetilde{T}}$ and $\mathcal{T}'$, the error we would have obtained had our learned $f$ \emph{exactly matched} the ground truth $\widetilde{f}$.  Because (\ref{eqn:sde}) is stochastic, this latter error will never be zero.

Carrying this out for $1000$ trajectories, we find that for $25$ trajectories, we obtain overflow errors from the ground truth polynomial vector field $\widetilde{f}$ in the $\alpha=1$ (Cauchy noise) case.  This does not happen with the Fourier representation $f$; we conjecture this is because $\widetilde{f}$ is not Lipschitz while $f$ is.  Eliminating the $25$ trajectories with overflow errors, we obtain a MMAE of $0.910$ and MIQR of $0.912$ for the comparison between $\widetilde{\mathcal{T}}$ and $\mathcal{T}$.  This is not far from the MMAE of $0.723$ and MIQR of $0.800$ for the comparison between $\widetilde{\mathcal{T}}$ and $\mathcal{T}'$.

%\subsection{Analyzing the Error for the Two-Dimensional Polynomial Double Well}
%\label{sect:twodimdblwell}

\end{document}